# Hazard-Responsive Digital Twin for Climate-Driven Urban Resilience and Equity

Zhenglai Shen[*,1], Hongyu Zhou[2]

[1] Buildings and Transportation Science Division, Oak Ridge National Laboratory, Oak Ridge, TN, 37831, USA
[2] Civil and Environmental Engineering, University of Tennessee, Knoxville, TN, 39702, USA

**Abstract**

Compounding climate hazards, such as wildfire-induced outages and urban heatwaves, challenge the stability and equity of cities. We present a Hazard-Responsive Digital Twin (H-RDT) that combines physics-informed neural network modeling, multimodal data fusion, and equity-aware risk analytics for urban-scale response. In a synthetic district with diverse building archetypes and populations, a simulated wildfire–outage–heatwave cascade shows that H-RDT maintains stable indoor temperature predictions (~31–33°C peaks) under partial sensor loss, reproducing outage-driven surges and recovery. The reinforcement learning based fusion module adaptively reweights IoT, UAV, and satellite inputs to sustain spatiotemporal coverage, while the equity-adjusted mapping isolates high-vulnerability clusters (schools, clinics, low-income housing). Prospective interventions, i.e., preemptive cooling-center activation and microgrid sharing, reduce population-weighted thermal risk by 11–13%, shrink the 95th-percentile (tail) risk by 7–17%, and cut overheating hours by up to 9%. Beyond the synthetic demonstration, the framework establishes a transferable foundation for real-city implementation, linking physical hazard modeling with social equity and decision intelligence. The H-RDT advances digital urban resilience toward adaptive, learning-based, and equity-centered decision support for climate adaptation.

[*] Corresponding Author: Email: shenz1@ornl.gov, TEL: 01-865-241-8772;
One Bethel Valley Road, P.O. Box 2008 MS-6070, Oak Ridge, TN 37831-6070, USA

# 1. Introduction

## 1.1 Background and Motivation

The increasing frequency and severity of climate-related disasters are reshaping the way societies perceive and manage risk. Complex events such as wildfires, floods, and heatwaves are no longer isolated phenomena but interlinked hazards that propagate through interconnected infrastructure networks. When one system fails, others that depend on it often cascade toward collapse, producing widespread disruption and social inequity. Recent crises including the 2023 Vermont flooding, the 2024 Texas winter freeze, and the 2025 Southern California wildfire illustrate how climate-amplified events can simultaneously strain energy, water, communication, and transportation systems. Traditional risk assessments, which often treat hazards as discrete and static events, are insufficient to capture the evolving and compounding nature of modern disasters.

Digital Twin (DT) technology offers a promising avenue for improving situational awareness and decision-making under such conditions. Originally introduced for aerospace engineering and later adopted across industrial sectors, DTs create real-time virtual counterparts of physical systems using sensor data, predictive modeling, and feedback control (Grieves & Vickers, 2018; Tao et al., 2019). Within the built environment, DTs have been applied to asset monitoring, predictive maintenance, and urban system management (Errandonea et al., 2020; Fogli, 2019; Fuller et al., 2020). However, most conventional DTs rely on stable connectivity, complete datasets, and deterministic control assumptions that are not held during crises characterized by cascading failures and data disruption.

To address these challenges, the concept of the Risk-Informed Digital Twin (RDT) integrates probabilistic modeling, uncertainty quantification, and decision support within the DT architecture (Pignatta & Alibrandi, 2022; Zio & Miqueles, 2024). Likewise, Digital Risk Twins (DRTs) combine human oversight and hybrid data integration to enable adaptive reasoning when information is incomplete or uncertain (Ghaffarian, 2025; Lagap & Ghaffarian, 2024a). Recent advancements in physics-informed digital twins (PIDTs) and hybrid frameworks have merged data-driven learning with physical constraints, improving reliability in noisy or sparse data environments (Karniadakis et al., 2021; H. Liang et al., 2025; Raissi et al., 2019; Jian Zhou et al., 2024). Despite this progress, existing digital twin systems often remain fragmented, focusing on single assets or isolated hazards rather than encompassing community-scale, multi-hazard resilience.

## 1.2 Literature Review

DT technology has evolved rapidly across disciplines. Early developments in manufacturing and Industry 4.0 enabled predictive control and real-time optimization (Kritzinger et al., 2018; Tao et al., 2019). In recent years, the concept has expanded into urban management through Urban Digital Twins (UDTs) and Digital Twin Cities (DTCs) that integrate geographic information systems (GIS), building information modeling (BIM), and Internet of Things (IoT) data (Ariyachandra & Wedawatta, 2023; Deren et al., 2021; Zhu & Jin, 2025). Examples such as Virtual Singapore and the UK's National Digital Twin Program demonstrate the potential of multi-domain data integration for energy management, mobility, and climate adaptation (Cimellaro et al., 2016).

Participatory and community-based models are extending these applications to social dimensions. Ham and Kum (Ham & Kum, 2020) developed a participatory sensing framework that allows citizens to contribute localized data to update urban digital twin models for risk-informed decision-making. Fan et al. (Fan et al., 2021) proposed a "disaster city digital twin" that combines artificial and human intelligence to improve disaster response coordination. These participatory efforts represent a shift from static data visualization toward adaptive, user-informed resilience platforms that integrate social and physical intelligence.

In parallel, hybrid and physics-informed twin architectures have been introduced to capture dynamic system behaviors under uncertainty. Tezzele et al. (Tezzele et al., 2024) presented adaptive planning for risk-aware predictive twins, while Liang et al. (H. Liang et al., 2025) proposed hybrid digital twinning for decision support in smart infrastructures. Zhou (Junjie Zhou, 2025) applied physics-informed machine learning to hybrid digital twins for enhanced damage detection and localization. Martini et al. (Martini et al., 2025) developed a distributed co-simulation approach to represent socio-technical systems for crisis management, and Cui et al. (Cui et al., 2024) introduced a digital twin platform for cryospheric disaster warning, demonstrating the versatility of such systems for environmental resilience.

Additionally, understanding interdependencies among infrastructure networks is central to resilience-oriented digital modeling. Foundational research by Rinaldi et al. (Rinaldi et al., 2001), Buldyrev et al. (Buldyrev et al., 2010), and Ouyang (Ouyang, 2014) showed how interdependent networks amplify cascading failures. Recent work has adopted graph-theoretic and machine-learning approaches to quantify these dependencies and identify critical vulnerabilities (Kritzinger et al., 2018; Wang et al., 2024; Jian Zhou et al., 2024). Liu and Meidani (Liu & Meidani, 2024) demonstrated how graph neural networks can capture evolving topologies within infrastructure systems, supporting dynamic risk assessment. Such advances underscore the value of combining physical models, network science, and artificial intelligence for predictive resilience.

On the other hand, resilience frameworks provide essential context for evaluating the adaptive performance of socio-technical systems. Bruneau et al. (Bruneau et al., 2003) defined resilience as the capacity to absorb, recover, and adapt following disruption. Compound hazard research further emphasizes that modern disasters are interdependent and sequential, producing overlapping social and environmental impacts (National Academies of Sciences, Engineering, and Medicine, 2024; Zscheischler et al., 2020). Multi-hazard digital twin frameworks, including the landscape-scale models of Ugliotti et al. (Ugliotti et al., 2023), the urban resilience models of Wang et al. (Wang et al., 2024), and power system frameworks by Braik and Koliou (Braik & Koliou, 2024), demonstrate the feasibility of integrating sensing, modeling, and control for complex hazard scenarios.

Moreover, human and equity dimensions are increasingly recognized as integral to resilience-oriented digital systems. Participatory sensing and volunteered geographic information (Goodchild, 2007; Kryvasheyeu et al., 2016) have improved situational awareness when traditional sensors fail, though they introduce challenges related to data quality and representativeness. Agent-based models complement digital twins by simulating human mobility, evacuation, and behavioral dynamics (Hawe et al., 2012; Sun et al., 2021). At a broader level, social vulnerability and energy justice research highlights that resilience measures often benefit populations unequally (Hamdanieh et al., 2024; Shen et al., 2022; Soden et al., 2023). Integrating socioeconomic indicators, demographic data, and participatory feedback within digital frameworks can improve equity and transparency in resilience planning.

Together, these studies highlight the evolution of DTs from asset-centric monitoring tools to hybrid, data-rich systems for dynamic urban resilience. However, existing implementations rarely unite physics-informed learning, network-based interdependency discovery, and equity-aware reasoning within a single adaptive framework capable of real-time operation under hazard conditions.

**1.3 Research Gaps and Objectives**

Despite significant advances in DT and DRT research, most current systems remain fragmented and domain-specific. They are typically designed for individual assets such as bridges, power grids, or water networks and thus fail to capture the cross-sector dependencies and multi-hazard interactions that characterize real communities. Three persistent limitations continue to restrict their practical resilience value: (1) communication fragility that disrupts synchronization between the physical and digital layers, (2) incomplete or inconsistent data that reduce model fidelity, and (3) limited integration of human judgment and ethical reasoning in automated processes. On the other hand, the DRT concept (Ghaffarian, 2025) provides an important bridge between data analytics and human expertise by incorporating hybrid sensing, remote observations, and manual oversight. While DRTs have improved situational awareness for applications such as flood management (Lagap & Ghaffarian, 2024b), power restoration (Braik & Koliou, 2024), and bridge maintenance (Geng et al., 2025), they remain narrowly scoped and do not yet support multi-hazard, cross-domain reasoning.

To address these gaps, this study introduces the Hazard-Responsive Digital Twin (H-RDT), a unifying framework that integrates sensing, physics-informed modeling, probabilistic reasoning, and participatory intelligence into a continuously evolving digital ecosystem. Unlike deterministic twins that assume steady-state operation, the H-RDT embraces uncertainty as an inherent feature of complex systems. It learns evolving interdependencies among hazards, infrastructure, and communities, and augments human decision-making rather than replacing it. Conceptually, the H-RDT envisions the digital twin as a cognitive collaborator that advances digital resilience toward anticipatory, adaptive, and equitable urban systems for a climate-challenged future.

## 2. The H-RDT
## 2.1 Architecture

The H-RDT is structured as an adaptive, multi-layer cyber-physical ecosystem that integrates environmental sensing, computational intelligence, and human decision-making into a continuously evolving feedback loop of perception, learning, and intervention. The architecture comprises three synergistic layers (Figure 1). It includes the physical layer, the digital-intelligence layer, and the decision layer.

The physical layer embodies real-world entities, including buildings, transportation networks, utilities, and populations, instrumented through a distributed network of fixed and mobile sensors. It assimilates heterogeneous data sources including ground sensors, unmanned aerial vehicles (UAVs), and satellite observations. When conventional infrastructure sensing is impaired, the layer incorporates ad-hoc and crowd-sourced information from social sensing and volunteered geographic data to preserve situational awareness. The digital-intelligence layer forms the computational core of the H-RDT. It fuses physics-informed neural networks (PINNs), finite-element surrogates, and reinforcement-learning (RL) modules to represent and continuously update system dynamics. These hybrid models learn the evolving dependencies among assets and hazards, capturing both physical causality and data-driven correlations. The integration of physics constraints within learning algorithms ensures that extrapolations remain consistent with thermodynamic, structural, or hydrological principles, while the RL component adapts control policies under nonstationary and compound hazard conditions. The decision layer provides human–machine interfaces for visualization, interpretation, and action. It renders probabilistic forecasts, uncertainty maps, and social-equity indicators to inform planners and emergency managers. Scenario-based simulations support both short-term operational response and long-term infrastructure planning by revealing trade-offs between resilience enhancement, cost, and social equity.

Data flow across the H-RDT is inherently bidirectional. Physical observations continuously update the digital state, while model forecasts inform control actions, policy decisions, and adaptive resource allocation. The coupled evolution of the system is expressed through a generalized state-space formulation:

$$\dot{\mathbf{x}} = \mathbf{A}\mathbf{x} + \mathbf{B}\mathbf{u} + \boldsymbol{\xi}(t) \tag{1}$$

$$\mathbf{y} = \mathbf{C}\mathbf{x} + \boldsymbol{\epsilon}(t) \tag{2}$$

where, **x** denotes the system state (e.g., temperatures, stresses, or service levels), **u** represents hazard or control inputs, and **y** is the observation vector from sensors. The stochastic terms $\boldsymbol{\xi}(t)$ and $\boldsymbol{\epsilon}(t)$ describe process and measurement uncertainties, respectively.

Real-time data assimilation is achieved through Kalman filtering or ensemble-based update schemes, enabling the digital twin to remain dynamically aligned with the physical environment. This continuous synchronization ensures that predictions, alerts, and decision recommendations produced by the H-RDT reflect the latest system conditions, thereby supporting proactive and equitable resilience management.

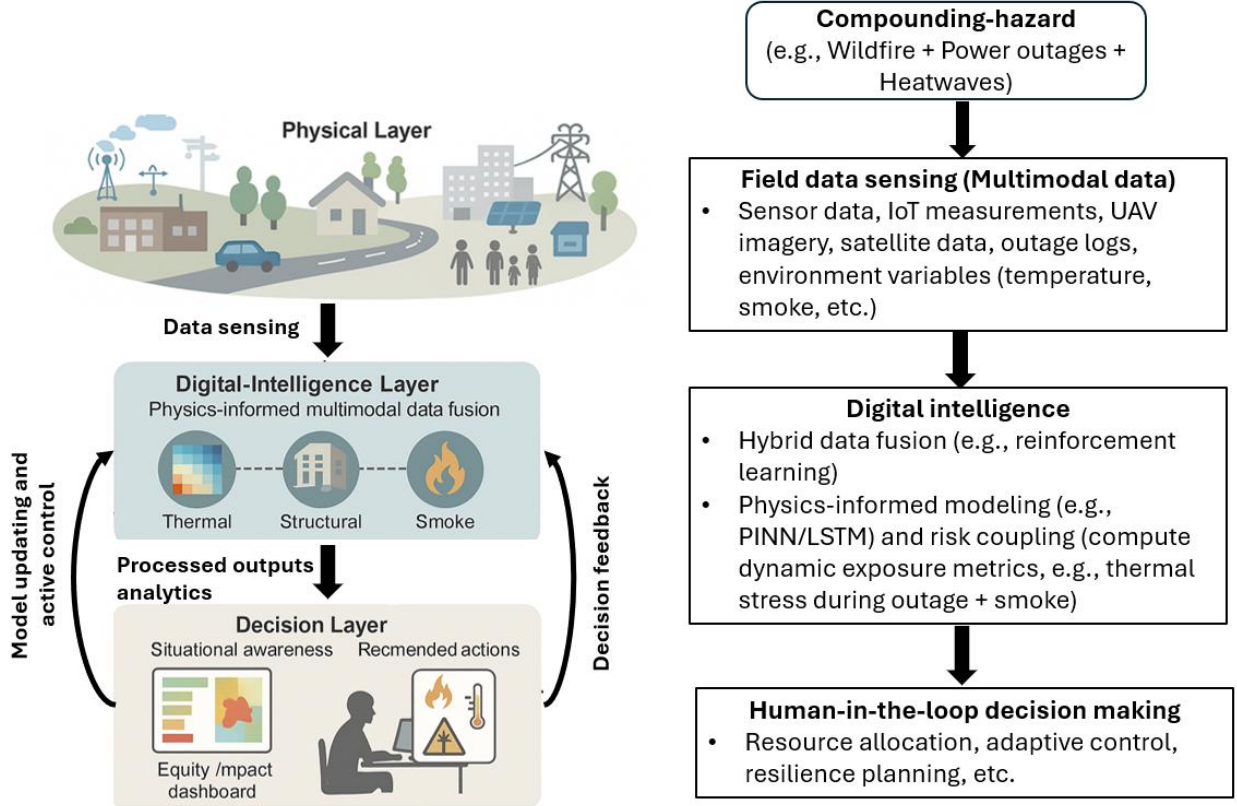

**Figure 1**. Conceptual overview of H-RDT: physical, digital-intelligence, and decision layers linked by continuous hybrid data exchange.

**2.2 Compounding-hazard characterization**

Unlike conventional risk assessments that treat hazards independently, the H-RDT conceptualizes them as dynamically coupled processes that evolve and interact across temporal and spatial scales. In this framework, hazards are not independent shocks but components of an interconnected system. For example, a wildfire disrupting electrical grids can exacerbate heat stress and indoor overheating; a flood may sever transportation routes and delay emergency response; and an earthquake can initiate cascading failures such as fires, gas leaks, or structural instability in its aftermath. These interdependencies are mathematically represented through a coupled hazard formulation (He & Weng, 2021; Tilloy et al., 2019):

$$H_c = \sum_j \omega_j f_j(H_j, T, S) \tag{3}$$

where $H_c$ denotes the compounding hazard stress, $H_j$ represents the intensity of individual hazards (e.g., seismic, thermal, or hydrologic), $T$ and $S$ correspond to temperature and structural states, and $\omega_j$ are adaptive importance weights dynamically updated via reinforcement learning to capture evolving hazard relevance.

The probability of one hazard triggering another is modeled through a transition matrix that encodes conditional dependencies:

$$P_{ij} = \Pr(H_j|H_i, T_i, S_i, t) = \psi(\theta_0 + \theta_1 T_i + \theta_2 S_i + \theta_3 t) \tag{4}$$

where $P_{ij}$ expresses the likelihood that hazard $H_i$ induces $H_j$ given the current temperature condition $T_i$, structural condition $S_i$, and elapsed time $t$; $\psi(\cdot)$ denotes a logistic activation ensuring probabilistic normalization, $\theta_0 - \theta_3$ are model coefficients that quantify the relative importance of temperature, structural vulnerability, and time in determining how one hazard triggers another. This stochastic representation enables H-RDT to infer both simultaneous and sequential hazard cascades, as well as delayed or amplifying effects that unfold over time.

The coupled hazard interactions ultimately reflected in the system's resilience function, which quantifies degradation and recovery dynamics (Gill & Malamud, 2014):

$$\mathcal{R}(t) = 1 - \frac{\int_{t_0}^{t}|L(t') - L_0|dt'}{L_0(t-t_0)} \tag{5}$$

where $L(t)$ denotes a performance metric such as service continuity, energy supply, or accessibility, and $\mathcal{R}(t)$ measures the normalized loss and restoration of functionality. Together, these coupled formulations establish the stochastic backbone of H-RDT's multi-hazard modeling framework, allowing it to capture how compounding hazards propagate, amplify, and reshape resilience trajectories across interconnected infrastructure and communities.

## 2.3 Hybrid data fusion

Effective data integration under uncertainty is a central function of the H-RDT framework. The system continuously assimilates heterogeneous data streams that differ in reliability, latency, and spatial coverage, encompassing automated, semi-automated, and human-derived sources. Automated data originates from Internet of Things (IoT) sensors, satellite radiometry, and LiDAR-derived topographic models, providing high-frequency, quantitative measurements of environmental and infrastructure conditions. Semi-automated data are collected through supervisory control and data acquisition (SCADA) systems, mobile inspection platforms, and drone-based reconnaissance, bridging local-scale detail with broader situational awareness. Human-derived data, obtained through social media, city information service (e.g., 311 service in the U.S.), or community surveys, fill observational gaps when formal sensing networks degrade or fail (Kryvasheyeu et al., 2016), contributing essential contextual knowledge about impacts and needs on the ground.

To reconcile these disparate inputs, each data stream $\varsigma$ is assigned a reliability weight $w_\varsigma$ that evolves adaptively through RL based on its historical performance:

$$w_\varsigma^{(t+1)} = w_\varsigma^{(t)} + \alpha\left(r_\varsigma^{(t)} - \bar{r}^{(t)}\right)\left(x_\varsigma^{(t)} - \bar{x}^{(t)}\right) \qquad (6)$$

where $r_\varsigma^{(t)}$ represents the reward signal measuring accuracy or internal consistency, $x_\varsigma^{(t)}$ denotes the assessed data quality at time $t$, $\alpha$ is the learning rate, and overbars indicate ensemble means across all sources. This update mechanism enables the H-RDT to down-weight unreliable or noisy sensors while increasing trust in verified or high-quality data streams, such as UAV-based observations or cross-validated crowdsourced inputs. Through this adaptive weighting, the H-RDT maintains robustness even under partial observability, i.e., redistributing trust dynamically across data modalities as environmental conditions and sensing reliability evolve. This hybrid data fusion process ensures that decision-support outputs remain grounded in the most credible and up-to-date information available.

**2.4 Physics-informed modeling and risk coupling**

At the building scale, the H-RDT integrates physics-based and data-driven models to represent the coupled thermal, structural, and energy dynamics governing hazard response. The physical core is grounded in conservation-based differential equations that describe how buildings exchange, store, and dissipate energy under evolving environmental and hazard conditions.

The indoor thermal environment of a building can be represented by a two-resistance, two-capacitance (2R2C) model that distinguishes between the thermal dynamics of the envelope (wall) and the indoor air zone. The coupled energy balance equations account for conductive and convective heat transfer, solar and internal gains, and the transient interaction between the wall temperature $T_w$ and indoor air temperature $T_z$ (Matthiss et al., 2023):

$$\begin{cases} C_w \frac{dT_w}{dt} = \frac{T_{out}-T_w}{R_{wo}} + \frac{T_z-T_w}{R_{wz}} \\ C_z \frac{dT_z}{dt} = \frac{T_w-T_z}{R_{wz}} + Q_{int} + Q_{sol} + Q_{hvac} \end{cases} \quad (7)$$

where $C_w$ and $C_z$ are the effective thermal capacitances of the wall and indoor air, $R_{wo}$ and $R_{wz}$ represent the thermal resistances between the outdoor air–wall and wall–zone interfaces, respectively, and $Q_{int}$, $Q_{sol}$, and $Q_{hvac}$ denote internal, solar, and HVAC heat gains. During power outages, $Q_{hvac}$ becomes zero, while the effective envelope conductance $1/R_{wo}$ may increase due to unconditioned infiltration and reduced ventilation control. Under wildfire smoke conditions, ventilation efficiency is reduced, which can be modeled by scaling $R_{wo}$ with a smoke modulation factor. This 2R2C formulation captures the distinct thermal lag between the building envelope and the indoor zone, providing a physically interpretable framework to evaluate heat retention, decay, and discomfort evolution during compounding outage–heatwave–smoke events.

For high-temperature or wildfire events, indoor conditions are additionally influenced by air and pollutant transport, which cannot be described by conduction alone. To capture these coupled effects, the H-RDT extends the lumped energy balance to a reduced-order advection–diffusion–generation framework (Drysdale, 2011):

$$\rho c_p \left(\frac{\partial T}{\partial t} + \mathbf{u} \cdot \nabla T\right) = \nabla \cdot (k \nabla T) + \dot{q}_{gen} \tag{8}$$

where $T$ is the air temperature, $\mathbf{u}$ the buoyancy-driven airflow, $k$ the thermal conductivity, $\rho$ the density, $c_p$ the specific heat, and $\dot{q}_{gen}$ the volumetric heat generation rate from equipment overheating or fire-induced combustion. This formulation extends classical Fourier heat conduction to include convective energy transport within enclosure spaces and is consistent with transient heat transfer treatments for high-temperature structural and compartment environments. At the envelope boundaries, the surface heat flux is governed by the combined convective–radiative boundary condition:

$$-k \frac{\partial T}{\partial n}\bigg|_\Gamma = h(T_s - T_\infty) + \varepsilon \sigma (T_s^4 - T_{sur}^4) \tag{9}$$

where $h$ is the convective heat-transfer coefficient, $\varepsilon$ the surface emissivity, and $\sigma$ the Stefan–Boltzmann constant. This representation aligns with established fire-exposure models for building elements that account for both radiative and convective exchanges at external and internal surfaces.

Simultaneously, indoor air quality (IAQ) becomes a critical component of hazard response, especially during wildfire smoke intrusion or ventilation loss. The H-RDT therefore augments the thermal model with a single-zone particulate balance (Fu et al., 2022):

$$\frac{dC^{PM2.5}}{dt} = a(t)PC_{out}^{PM2.5} - \left[a(t) + k_{dep} + \frac{CADR}{V}\right]C^{PM2.5} + S_{ind} \qquad (10)$$

where $C^{PM2.5}$ is the indoor PM2.5 concentration, $C_{out}^{PM2.5}$ the outdoor concentration, $a(t)$ the air-exchange rate (ACH/3600), $P$ the envelope penetration factor, $k_{dep}$ the deposition rate, CADR the clean-air delivery rate of any portable air cleaner, $V$ the zone volume, and $S_{ind}$ indoor emission sources. This balance form is widely employed to evaluate indoor exposure to wildfire smoke and to quantify pollutant infiltration and removal (Y. Liang et al., 2021).

The structural response of the building under dynamic or sequential hazards is captured through a simplified damage evolution law that relates stress or strain history to material degradation (Lemaitre, 1985):

$$D(t) = 1 - e^{-\beta(\varepsilon s(t) - \varepsilon s_0)}, \quad K_{eff} = (1-D)K_0 \qquad (11)$$

where $D(t)$ is the damage index, $\varepsilon s(t)$ represents the strain history, $\beta$ is the material sensitivity coefficient, and $K_{eff}$ is the effective degraded stiffness. This exponential-type softening law has been widely used in continuum damage mechanics (CDM) and thermomechanical degradation models to describe progressive stiffness loss in concrete and quasi-brittle materials. It allows H-RDT to track cumulative damage from thermal stress, seismic motion, or impact loading and to infer post-event structural capacity.

All physical models within the H-RDT are implemented through a Physics-Informed Neural Network (PINN) framework, which constrains machine-learning predictions to satisfy the governing conservation laws of energy, momentum, and material behavior. Unlike conventional neural networks that learn solely from data correlations, the PINN embeds the underlying physics, such as transient heat conduction, structural equilibrium, and damage evolution, directly into its optimization process. The total loss function balances empirical data fidelity with physical consistency (Raissi et al., 2019):

$$\mathcal{L}_{\text{PINN}} = \lambda_{\text{data}}\|\hat{y} - y_{\text{obs}}\|^2 + \lambda_{\text{phys}}\|\aleph|\hat{y}|\|^2 \tag{12}$$

where $\hat{y}$ denotes the network-predicted field variable (e.g., temperature, displacement, or damage index), $y_{\text{obs}}$ represents the corresponding measured data, and $\aleph|\hat{y}|$ is the residual of the governing partial differential equation. The weighting coefficients $\lambda_{\text{data}}$ and $\lambda_{\text{phys}}$ control the trade-off between fitting available observations and enforcing physical constraints.

By minimizing both data and physics residuals simultaneously, the PINN produces spatial-temporal fields that remain faithful to sensor information while inherently obeying conservation principles. This coupling of thermal, structural, and stochastic risk representations enables the H-RDT to capture the full spectrum of hazard interactions at the building levels supporting reliable simulations of cascading failures, recovery trajectories, and adaptive responses under uncertainty.

**2.5 Learning affiliation networks**

Urban resilience emerges from evolving functional interdependencies, where schools temporarily serve as shelters, grocery stores repurposed as logistics hubs, or microgrids sustaining critical loads during outages. Within the H-RDT, these evolving roles are captured through a graph reinforcement-learning (GRL) module that dynamically updates the affiliation network among community assets.

Let $\mathbf{A}_t$ denote the adjacent matrix describing link strengths among $N$ assets at time $t$. The update rule follows a gradient-based reinforcement step:

$$\mathbf{A}_{t+1} = \mathbf{A}_t + \eta \nabla_{\mathbf{A}} \mathcal{L}_{RL}(\mathbf{A}_t, \mathbf{r}_t) \tag{13}$$

where $\eta$ is the learning rate and $\mathcal{L}_{RL}$ is the loss derived from observed reward vectors $\mathbf{r}_t$ representing system-level performance (e.g., accessibility, service continuity, temperature stability).

During simulation, the GRL algorithm evaluates how each node's operation contributes to community outcomes under stress, such as reduced overheating hours or improved accessibility during rolling outages, meanwhile incrementally adjusts edge weights to reinforce beneficial dependencies. Through this adaptive process, the network autonomously identifies emergent critical nodes, i.e., facilities whose degradation disproportionately impacts neighboring systems, and reveals latent pathways of mutual support. These learned affiliations provide actionable insights for prioritizing energy sharing, emergency services, or infrastructure reinforcement within the broader resilience strategy (Wu et al., 2021).

**2.6 Equity-adjusted risk integration**

While the current H-RDT implementation does not yet include a full human-in-the-loop interface, it embeds equity directly into quantitative risk estimation through a two-tier formulation encompassing both node (building) scale and urban-scale indices.

At the node scale, equity-adjusted risk ($R_{node}$) is computed as:

$$R_{node} = \text{exposure}_{sys} \times (\beta_{phys}V_{rank} + \beta_{sens}E_{rank}) \tag{14}$$

where $\text{exposure}_{sys}$ quantifies concurrent hazard stress (e.g., outage and thermal exposure), $V_{rank}$ and $E_{rank}$ are the percentile-ranked physical vulnerability and socio-economic sensitivity of each building or population cluster, and $\beta_{phys}$ and $\beta_{sens}$ are weighting coefficients reflecting their relative importance (0.6 and 0.4, respectively).

At the urban scale, these node scale values are aggregated to from the equity-adjusted urban scale risk index:

$$R_{eq} = \frac{\sum_k P_k V_k [1+\gamma E_k]}{\sum_k P_k} \tag{15}$$

where $P_k$ is the population associated with node $k$, and $\gamma$ a scaling coefficient that amplifies the contribution of disadvantaged populations to the total system risk. This hierarchical structure ensures that the spatial aggregation of risk preserves social weighting, giving proportionally greater emphasis to communities with higher vulnerability or lower adaptive capacity.

The decision layer visualizes spatial gradients of $R_{eq}$ to highlight neighborhoods where physical exposure and social vulnerability intersect most severely. These maps enable planners to assess counterfactual scenarios, such as targeted microgrid deployment or pre-emptive activation of cooling-center. They also establish a quantitative link between physical resilience modeling and social-equity objectives. Future development will extend this component toward participatory validation, allowing community stakeholders and local experts to review simulation outcomes through interactive dashboards. This pathway will establish genuine human-in-the-loop integration which combines quantitative rigor with lived experience to refine parameters and ensure contextual legitimacy.

**2.7 Relief and Intervention Modeling**

The H-RDT framework extends beyond passive risk assessment to simulate and evaluate diverse relief and intervention strategies that mitigate multi-hazard impacts and accelerate community recovery. This transforms the digital twin from a diagnostic tool into an adaptive decision-support platform, enabling planners to dynamically test, compare, and prioritize interventions under evolving hazard conditions. Each strategy modifies one or more parameters of the node-level equity-adjusted risk formulation (**Section 2.6**), adjusting exposure, physical vulnerability, or socio-economic sensitivity.

At the physical level, interventions focus on exposure reduction and infrastructure protection. The specific mechanisms depend on the dominant hazard type. For thermal and outage hazards, measures such as distributed microgrids, reflective roofs, and mobile cooling systems reduce both the frequency and duration of thermal stress. For seismic hazards, base isolators and structural retrofits lower structural fragility and prevent collapse. For flood resilience, elevated foundations, flood barriers, and water-resistant materials decrease inundation exposure. For hurricane and wind hazards, aerodynamic façades, reinforced roofing, and storm shutters limit wind intrusion and envelope damage.

At the urban and community scale, social and institutional interventions are represented through modifications to vulnerability and sensitivity weights, capturing enhancements in preparedness, awareness, and response capacity. Examples include pre-event activation of public shelters, adaptive evacuation routing, targeted community outreach, distributed retrofit incentives, and post-disaster recovery aid. These adjustments alter the socio-economic sensitivity term in the equity function, reflecting improved access to resources and institutional support. The combined representation of physical and social adjustments enables the H-RDT to quantify both direct (engineering-based) and indirect (behavioral and policy-driven) pathways toward resilience improvement across multiple hazard types. In the H-RDT formulation, these physical measures are encoded as multiplicative reductions in hazard intensity or exposure probability, directly influencing the system's state variables governing energy balance, service continuity, and structural integrity. This representation allows the framework to trace how physical interventions modify both immediate hazard impacts and long-term recovery trajectories.

Each intervention alters one or more parameters of the equity-adjusted risk formulation, which combines (i) exposure: the physical intensity or duration of hazard stress, and two dimensionless indices: (ii) physical vulnerability ($V_{rank}$), representing the structural fragility or robustness of buildings, and (iii) socioeconomic sensitivity ($E_{rank}$), denoting the adaptive capacity of the occupants and their access to resources. Lowering either $V_{rank}$ or $E_{rank}$ indicates improved physical integrity or enhanced social preparedness, respectively, both of which reduce the overall equity-adjusted risk at each node.

To evaluate the effectiveness of each intervention, the H-RDT computes three complementary performance metrics that capture both average and distributional improvements in resilience: population-weighted risk reduction ($\Delta P_{pop}(\%)$), equity-tail improvement ($\Delta R_{95}(\%)$), and overheating-hour reduction ($\Delta OH(\%)$) that capture average and distributional improvements in resilience. The population-weighted risk reduction is defined as:

$$\Delta P_{pop}(\%) = 100 \times \frac{\frac{\sum_k p_k R'_k}{\sum_k p_k} - \frac{\sum_k p_k R_k}{\sum_k p_k}}{\frac{\sum_k p_k R_k}{\sum_k p_k}} \tag{16}$$

where, $R_k$ and $R'_k$ are the baseline and post-intervention equity-adjusted risks, and $p_k$ represents population or exposure weight for node $k$.

The equity-tail metric, denoted as $\Delta R_{95}(\%)$, measures improvement for the most vulnerable nodes:

$$\Delta R_{95}(\%) = 100 \times \frac{Q_{95}(R'_i) - Q_{95}(R_i)}{Q_{95}(R_i)} \tag{17}$$

where $Q_{95}(\cdot)$ is the 95th-percentile operator across all nodes. Negative values indicate improvement.

Finally, the overheating-hour change ($\Delta OH(\%)$), measures percentage variation in cumulative hours exceeding a predefined thermal threshold (e.g., 30°C) before and after each intervention:

$$\Delta OH(\%) = 100 \times \frac{OH' - OH}{OH} \tag{18}$$

This indicator reflects how effectively an intervention mitigates prolonged heat exposure during compounding outage-heat events.

To assess implementation efficiency, these metrics can be normalized by cost ($C$) or staffing demand ($S$) to define composite indicators such as $E_c = \Delta R_{pop}/C$, $E_S = \Delta R_{pop}/S$, which form Pareto-fronts identifying the most effective and resource-efficient measures. Together, these formulations establish a transparent and extensible foundation for multi-hazard relief planning. By embedding quantitative performance indicators directly within the simulation loop, the H-RDT continuously reprioritizes interventions as new data, forecasts, and equity objectives emerge which enables adaptive management and equitable recovery under future compound hazard scenarios.

## 3. Case Study: Wildfire-Outage-Heatwave Cascade

Compound wildfire–outage–heatwave events present one of the most complex frontiers in urban resilience analysis, as they simultaneously stress physical infrastructure, building microclimates, and social systems. To demonstrate the capability of H-RDT, we simulated a hypothetical event representing conditions typical of western U.S. metropolitan regions during late-summer wildfire seasons.

**3.1 Case study configuration and cascading hazard scenario**

A synthetic urban district comprising 120 buildings (**Figure 2**) was generated to evaluate the H-RDT framework under compounding wildfire–outage–heatwave stressors. The buildings are distributed uniformly within a unit square, each assigned one of six functional types, including multi-family, single-family, commercial, school, grocery, and clinic, according to the probability distribution listed in **Table 1**. To ensure that all critical facilities are represented, at least one school, two groceries, and one clinic are explicitly enforced in the generation process. Each building node $k$ is endowed with physical, socioeconomic, and sensing attributes reflecting its function and location. Populations are sampled from type-specific ranges (e.g., 20–60 occupants for multi-family housing, 100–500 for schools), while income increases spatially toward the upper-right quadrant of the district. Energy burden, inversely related to income, varies between 0 and 1 with mild spatial heterogeneity, and physical vulnerability is defined by type with Gaussian noise to capture intra-class variability. Approximately 10% of the nodes (12 buildings) are equipped with IoT temperature sensors subject to outage-induced dropouts. The computational experiment spans three days (72 hours) with a 10-minute time step, producing roughly 432 temporal samples per node. Outdoor temperature $T_{out}(t)$ follows a diurnal sinusoidal function representative of a heat-wave event, peaking near 40°C. Rolling power outages occur every six hours and persist for four hours per cycle, temporarily disabling cooling systems and selected sensors. **Figure 3(a)** presents the indoor and outdoor temperatures, including example indoor zone air temperature for a school and a clinic, and the HVAC setpoint. Wildfire smoke infiltration is introduced through a ventilation-modulation coefficient of 0.15, reducing the effective heat exchange between indoor and outdoor air. Latent indoor temperatures are modeled as building-specific couplings to $T_{out}(t)$ with random offsets, smoke-modulated diurnal factors, and a +1.2°C drift during blackout periods. The full set of temporal and hazard parameters is summarized in **Table 2**.

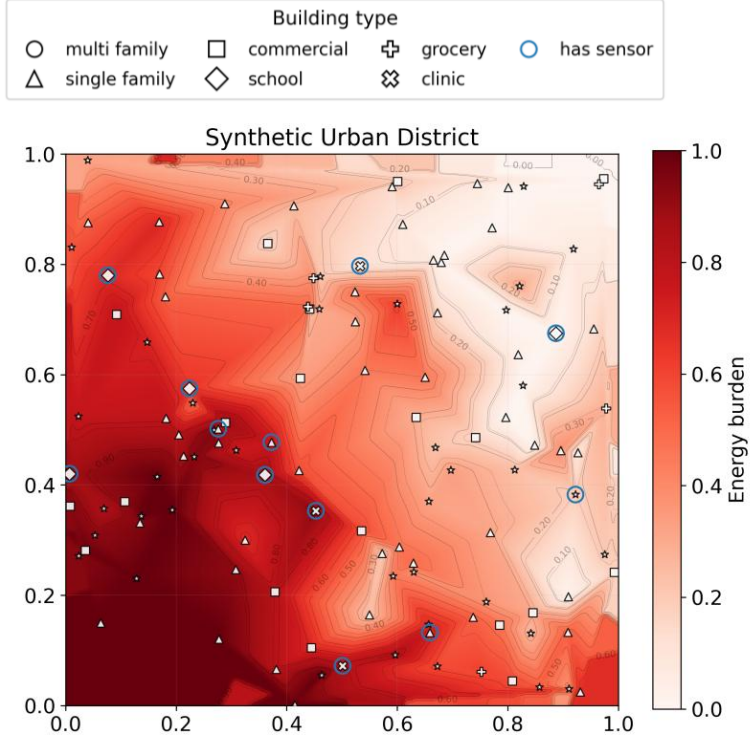

**Figure 2.** Synthetic urban district layout illustrating building types, sensor locations, and energy burden surface used for scenario simulations

Three asynchronous sensing modalities emulate realistic data acquisition during hazard-induced disruptions. IoT devices provide 10-minute measurements with ±0.4°C uncertainty and complete data loss during outages, UAV thermal imagery supplies hourly snapshots with ±0.8°C noise, and satellite radiometry contributes coarse observations every six hours with ±1.2°C uncertainty (**Table 2**). **Figure 3(b)** shows the representative indoor air temperature profile for the School, along with corresponding virtual measurements derived from IoT sensors, UAV observations, and satellite data. These heterogeneous sources feed the H-RDT's data-fusion module, which uses reinforcement-learning based adaptive weighting to maintain situational awareness when one or more streams fail. Together, the configuration in **Table 1** and the simulation protocol in **Table 2** reproduce the spatial heterogeneity, temporal intermittency, and data incompleteness typical of real-world multi-hazard scenarios, providing a testbed for assessing the robustness of the proposed digital-intelligence framework.

**Table 1.** Synthetic district configuration and node attributes

| Attribute | Symbol / Value | Description |
|---|---|---|
| Number of buildings | (N = 120) | Total building nodes in unit square |
| Building types (fractions) | 0.45 / 0.33 / 0.15 / 0.02 / 0.03 / 0.02 | Multi-family (MF) / Single-family(SF) / Commercial (COM) / School (SCH) / Grocery (GRO) / Clinic (CLI) |
| Population per node | $P_k$ | MF 20–60; SF 2–5; COM 1–15; SCH 100–500; GRO 20–80; CLI 10–60 |
| Income | $I_k \in [20k, 180k]$ | $30 + 70(0.5x_i + 0.5y_i) + N(0,8)$ |
| Energy burden | $E_k \in [0, 1]$ | Inversely related to income + spatial sinusoid |
| Physical vulnerability | $V_k \in [0, 1]$ | Baselines ± $N(0,0.05)$: MF 0.55; SF 0.45; COM 0.40; SCH 0.70; GRO 0.50; CLI 0.65 |
| Sensor availability | 12 nodes (10%) | IoT sensors with dropout during outages |

**Table 2.** Simulation and sensing parameters

| Parameter | Symbol / Value | Description |
|---|---|---|
| Simulation duration | 3 days | Total period simulated |
| Time step | 10 min | Temporal resolution (432 steps) |
| Outdoor temperature | $T_{out}(t) = 32 + 8\sin[2\pi/(t/24 - 0.15)]$ | Diurnal heat-wave cycle, peak ~40°C |
| Power outage | Period 6 h, length 4 h | Rolling outages across all nodes |
| Smoke modulation coefficient | 0.15 | Reduction in ventilation efficiency |
| Indoor temperature drift | +1.2°C | Rise during outage events |
| IoT sensors | 10-min sampling ± 0.4°C noise | Dropout during outage |
| UAV imagery | Hourly frames ± 0.8°C noise | Fine-grained thermal mapping |
| Satellite data | 6-h intervals ± 1.2°C noise | Coarse regional thermal field |
| Data fusion | RL-based adaptive weighting | Dynamically adjusts source confidence |

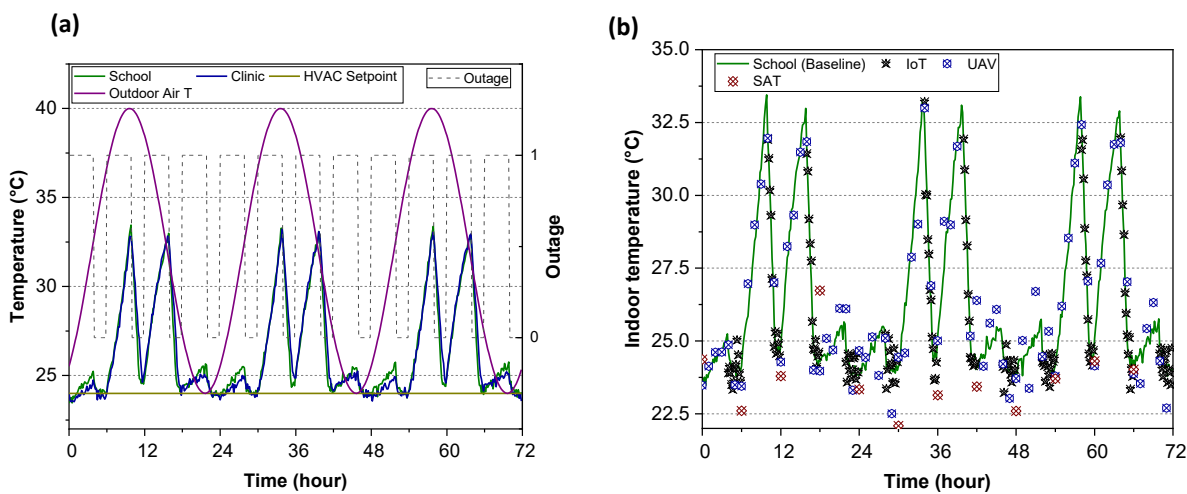

**Figure 3.** Indoor, outdoor, and measured temperatures: (a) Indoor and outdoor temperature (indoor zone air temperature including school and clinic, and HVAC setpoint); and (b) sensor measurements including baseline, IoT, UAV, and SAT (satellite).

### 3.2 Results and discussions

*3.2.1 PINN performance*

The PINN algorithm for indoor air temperature prediction under the compound wildfire–outage–heatwave is detailed in **Table A1** in the Appendix. **Figure 4** presents the PINN prediction using the 2R2C model (**Eq. (7)**) for the indoor air temperature over a 72-hour compounding hazard scenario encompassing a heatwave, power outage, and smoke exposure. The model effectively captures the diurnal temperature cycles and the outage-induced thermal surges across all buildings in the synthetic urban environment.

For both buildings, the PINN demonstrates strong agreement with measured data while preserving physical realism. In the School case (**Figure 4(a)**), the model precisely tracks the morning warm-up and reproduces sharp afternoon peaks of about 31–33 °C during blackout periods before gradually returning to baseline temperatures overnight as cooling resumes. The predicted peaks align with IoT measurements within a single sampling interval, confirming that the learned thermal parameters and the embedded outage–smoke interaction logic jointly capture realistic heat-transfer dynamics. Small nighttime residuals (0.3–0.6 °C) arise from unmodeled ventilation and sensor noise but remain bounded and non-cumulative. In the Clinic case (**Figure 4(b)**), the model reflects a greater effective thermal mass and the short-term oscillations associated with HVAC cycling prior to outages. Despite abrupt power transitions, the PINN maintains numerical stability, accurately reproducing the onset and recovery of outage-induced temperature surges. A modest cool bias occurs during prolonged outages, likely due to stronger real-world ventilation or internal gains than represented in the priors.

Quantitative validation results are summarized in **Table 3**, showing that the overall RMSE and MAE are 1.93 °C and 1.42 °C, respectively. IoT-driven predictions exhibit the best performance (RMSE = 1.60 °C, MAE = 1.22 °C), followed by UAV-derived data (RMSE = 1.99 °C, MAE = 1.46 °C). Satellite-based evaluation is excluded due to insufficient temporal coverage. Collectively, these results confirm that the PINN generalizes across building types, retains physical interpretability, and exhibits graceful degradation under sensing intermittency which is a critical capability for resilience assessment during compound hazard conditions.

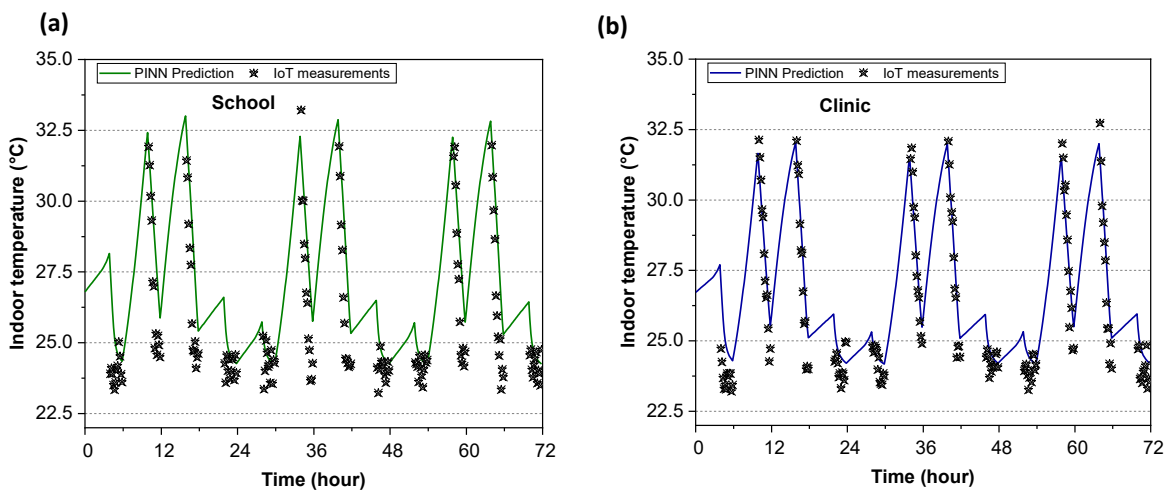

**Figure 4.** PINN model prediction vs. IoT measurements for: (a) School; and (b) Clinic.

**Table 3**. PINN validation metrics across multimodal data sources

| Data Source | *RMSE (°C) | **MAE (°C) |
|---|---|---|
| IoT Sensors | 1.6 | 1.22 |
| UAV Thermal | 1.99 | 1.46 |
| Satellite (SAT) | — | — |
| Overall (Pooled) | 1.93 | 1.42 |

*RMSE = Root Mean Square Error
**MAE = Mean Absolute Error

*3.2.2 Fusion weights of sensors*

The adaptive multimodal fusion algorithm for estimating the weights for IoT, UAV, and satellite is presented in **Table A2** in the Appendix. The adaptive fusion weights, as shown in **Figure 5**, illustrate the dynamic contribution of IoT, UAV, and satellite (SAT) data streams to the multimodal assimilation framework over the 72-hour compounding hazard event. Initially, the IoT data stream receives higher confidence due to its dense temporal coverage and low latency; however, as the system detects greater noise, missing values, or potential signal drift, its fusion weight gradually decreases from about 0.32 to 0.22. In contrast, the UAV and SAT streams progressively gain influence as they provide complementary, spatially aggregated observations that remain temporally fresh within their respective update intervals. The UAV weight stabilizes near 0.40, while the SAT weight oscillates around 0.35 with periodicity reflecting its lower sampling frequency and smoother temperature field. These adaptive trajectories confirm that the adopted Soft Maximum Function (softmax)–Exponential Moving Average (EMA) fusion mechanism (see **Table A2**) responds consistently to variations in data quality, availability, and freshness. By automatically re-balancing information sources, the algorithm prevents over-reliance on any single stream and sustains reliable state estimation even under sensing intermittency, which is an essential capability for maintaining situational awareness during multi-hazard and partial-connectivity scenarios.

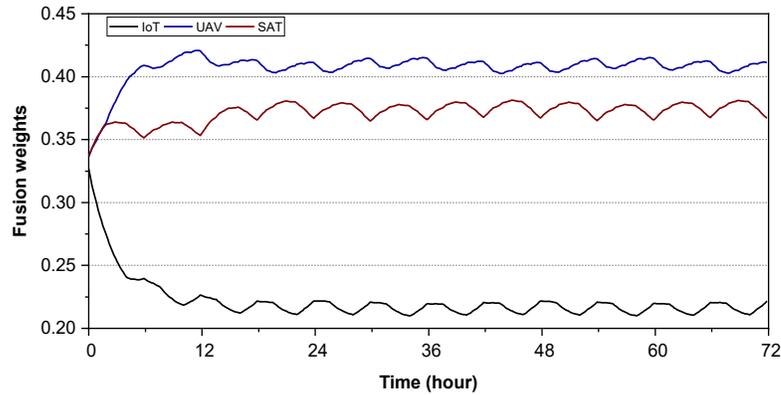

**Figure 5.** Fusion weights for IoT, UAV, and Satellite measurements.

*3.2.3  GRL update with coverage-aware critical and edge-gain overlay*

The GRL update with coverage-aware critical and edge-gain overlay algorithm is summarized in **Table A3** in the Appendix. The results are illustrated in **Figure 6**, which visualizes how inter-building connectivity evolves under compound stress conditions. **Figure 6(a)** shows the betweenness-based centrality heatmap after the GRL iteration. Nodes with deeper red colors represent facilities that have become increasingly influential in maintaining communication or service continuity across the urban network. These nodes act as bridge points which connect clusters of otherwise weakly linked communities. They tend to emerge around the central and western sections of the district, where population density and service dependency are highest. The surrounding lighter nodes indicate peripheral buildings or low-traffic links that play minor roles in network resilience. The dense connectivity and clustering visible in the graph demonstrate that the GRL algorithm successfully reinforces paths between spatially proximate yet functionally important facilities, producing a more coherent backbone for information and resource flow during hazard escalation.

**Figure 6(b)** presents the affiliation graph following the GRL update, highlighting the coverage-aware critical nodes in red. These nodes represent the top-ranking facilities within each community cluster, identified through a combination of centrality and demand metrics. They form an optimized balance between local coverage and global influence, ensuring that every sub-region retains at least one strategically connected hub even under partial network disruption. The spatial spread of these red nodes across the map suggests an equitable distribution of critical assets, avoiding over-concentration of resilience capacity in a single area. Compared with the baseline graph, the post-GRL configuration reflects a network that is both structurally robust and coverage-efficient which is ideal for guiding emergency coordination, microgrid activation, and data-driven decision-making during outages or heat-induced stress events.

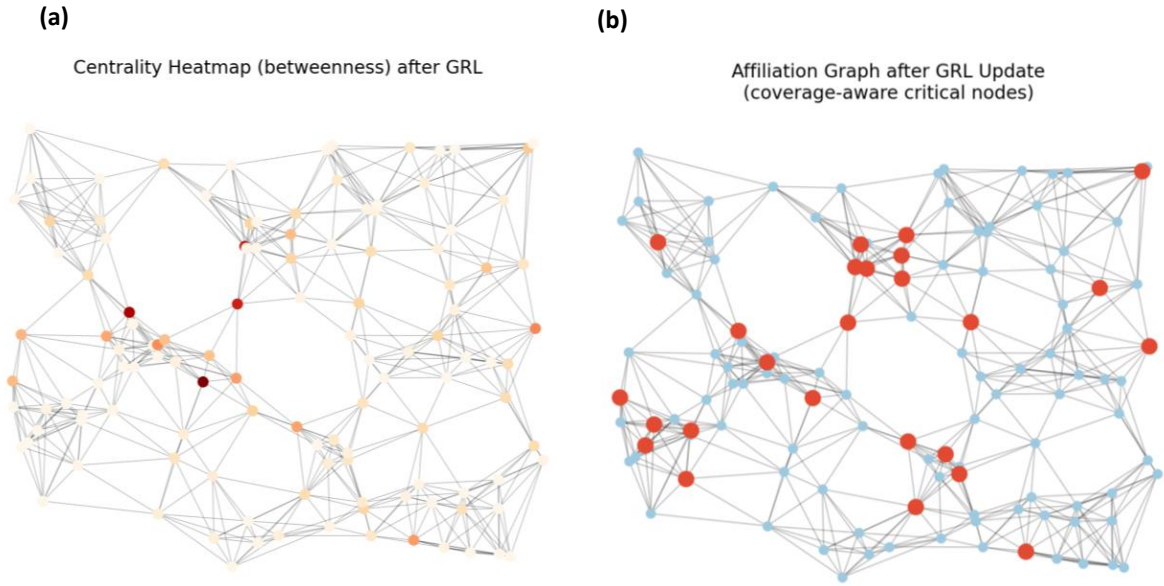

**Figure 6.** GRL enhances urban network resilience by: (a) strengthening high-betweenness corridors; and (b) identifying coverage-aware critical facilities.

*3.2.4 Equity-adjusted risks*

The equity-adjusted risk algorithm is summarized in **Table A4** in the Appendix. The equity-adjusted risk results are visualized in **Figure 7**, which compares the per-node (building-level) distribution and the smoothed regional surface of risk across the simulated urban environment. **Figure 7(a)** shows the discrete equity-adjusted risk for each node, revealing a heterogeneous pattern where certain clusters of buildings, particularly those concentrated in the lower and central portions of the map, exhibit higher combined vulnerability. These areas represent nodes with elevated physical fragility (e.g., schools, clinics, or low-resilience housing) coupled with greater social or energy burdens. In contrast, the upper-right quadrant contains several low-risk nodes, suggesting better infrastructure or socioeconomic conditions that buffer against compound hazards. The variation in shading from light to deep blue reflects how the model captures both inter-building disparities and the compounded influence of heat exposure, outage duration, and population sensitivity.

**Figure 7(b)** presents the smoothed equity-adjusted risk surface, which provides a continuous spatial interpretation of systemic inequity across the synthetic district. The spatial interpolation reveals several "risk corridors" where equity disparities align with local density and building typology, illustrating how compounded hazards can disproportionately affect socioeconomically or physically vulnerable zones. The smoother gradient emphasizes the influence of spatial proximity: clusters of high-risk buildings tend to reinforce neighboring risks through heat amplification and limited adaptive capacity. Overall, the results confirm that the combined weighting of physical vulnerability and social burden produces a realistic equity-sensitive hazard surface that can guide mitigation prioritization and targeted resilience planning within a simulated urban context.

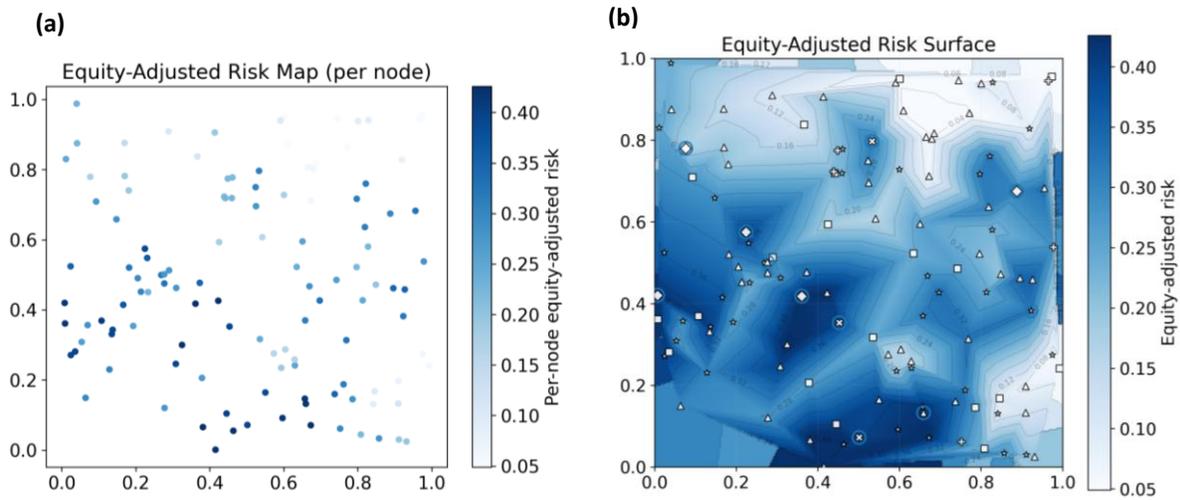

**Figure 7.** Equity adjusted risk maps for: (a) each building (per node); and the synthetic district

Complementing the spatial interpretation in **Figure 7**, **Figure 8** quantifies the distribution of equity-adjusted risk across population deciles, offering a clearer view of inequality in hazard exposure and adaptive capacity. The deciles (D1–D10) represent ranked groups of buildings or population clusters divided into ten equal portions based on their equity-adjusted risk scores, with D1 denoting the lowest-risk (most resilient) 10% of nodes and D10 indicating the highest-risk (most vulnerable) 10%. For each decile, the model computes both the arithmetic mean and the population-weighted average of the equity-adjusted risk, the latter emphasizing areas with higher population density or social sensitivity. As shown in **Figure 8**, both mean and population-weighted risks increase monotonically from D1 to D10, illustrating that a relatively small subset of the population bears a disproportionately large share of cumulative risk. The widening gap between the two metrics in the upper deciles indicates that highly populated or socially vulnerable clusters contribute more heavily to overall inequity, as these areas experience greater thermal exposure, longer outage durations, and reduced capacity for self-adaptation. This decile-based analysis highlights how the H-RDT framework captures not only spatial heterogeneity but also the population-weighted amplification of risk, enabling targeted policy interventions that prioritize high-burden communities and critical-service buildings.

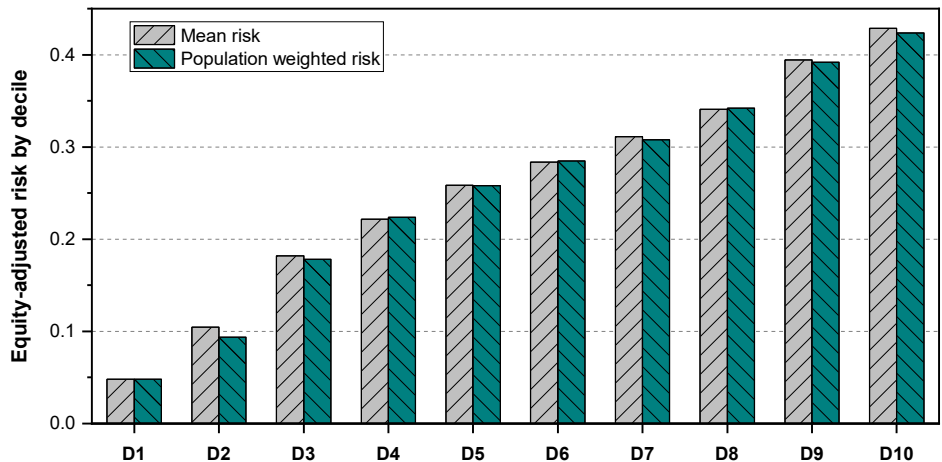

**Figure 8.** Equity adjusted risk by decile

*3.2.5 Intervention over equity-adjusted risks*

Five representative strategies were modeled to capture both infrastructural and urban-scale adaptation pathways, as summarized in **Table 4**, with implementation details provided in **Table A5** in the Appendix. The interventions outlined in **Table 4** include:

(1) **I1 – Preemptive Cooling Center**: Early opening of public shelters before the onset of heat or outage conditions, following evidence that anticipatory activation of cooling facilities can reduce heat exposure and associated morality by 25%–35% (Rohat et al., 2021; U.S. EPA, 2016). In this study, exposure was set to reduce by 30%, while vulnerability ($V_{rank}$) and sensitivity ($E_{rank}$) ranks decrease by 0.05 and 0.10, respectively, reflecting proactive preparedness and outreach.

(2) **I2 – Reactive Opening**: Activation of shelters only after critical thresholds are exceeded, consistent with observed reactive municipal responses that achieve limited benefits (~5%–10% reduction in exposure) once hazardous conditions emerge (Hondula et al., 2015; Petkova et al., 2017). We set the exposure reduced by 10%, and $E_{rank}$ decreases by 0.05, reflecting delayed institutional response.

(3) **I3 – Microgrid (Clinic-First)**: Localized power backup for healthcare facilities through distributed microgrids, which enhance reliability and ensure continuity of critical services during outages (Booth et al., 2019). Exposure was halved (× 0.5) and $V_{rank}$ reduced by 0.05 to represent strengthen infrastructure at critical nodes.

(4) **I4 – Microgrid (Vulnerable Block)**: Expansion of microgrid coverage to the top 15% of high-risk buildings, cutting exposure by 35%. This scenario reflects neighborhood-scale resilience investment that reduces outage duration and thermal burden for socially vulnerable groups (Hamidieh & Ghassemi, 2022; Panteli & Mancarella, 2017).

(5) **I5 – Targeted Outreach/Retrofits**: Community engagement and household retrofit programs that reduce $V_{rank}$ by 0.03 and $E_{rank}$ by 0.12, representing behavioral and financial support to improve household resilience. This is consistent with findings that retrofit incentives and behavioral intervention enhance energy equity and thermal resilience among low-income households (Reames, 2016; Tonn et al., 2014).

Each case was further assigned approximate staff hours per day and total implementation cost (×1000 USD) to reflect practical resource demands.

All interventions achieved measurable reductions in equity-adjusted risk across the simulated district, with performance visualized in **Figure 9**. The Microgrid – Vulnerable Block (I4) and Preemptive Cooling Center (I1) exhibited the strongest overall improvements, reducing the population-weighted risk ($\Delta P_{pop}(\%)$) by approximately 11%–13% and the 95th-percentile risk ($\Delta R_{95}(\%)$) by 7%–17%, and the cumulative overheating ($\Delta OH(\%)$) by 7%–9%. The Reactive Opening (I2) achieved a moderate benefit (~4% reduction in $\Delta P_{pop}(\%)$, 7% in $\Delta R_{95}(\%)$, and 2% in $\Delta OH(\%)$ ), which is consistent with its delayed activation and limited temporal coverage. In contrast, while Targeted Outreach (I5) and Microgrid Clinic-First (I3) yielded smaller but focused improvements (1%–3%), primarily benefiting socially vulnerable households or essential-service buildings. The differential performance reflects both spatial reach and timing of implementation. Preemptive measures (I1, I4) act before hazard escalation, effectively lowering systemic exposure and redistributing resilience benefits across a wider population base. Reactive or localized actions (I2–I3–I5) produce incremental yet equitable benefits for specific subpopulations.

The Pareto-front analysis (**Figure 9**) reveals that the Microgrid – Vulnerable Block (I4) intervention provides the highest efficiency per investment, achieving the greatest reduction in population-weighted (–12.9 %) and 95th-percentile (–16.9 %) risk with no recurring staffing demand. In contrast, the Preemptive Cooling Center (I1), while also Pareto-optimal, requires continuous daily staffing but achieves comparable reductions ($\approx$ –11.6 % in $\Delta P_{pop}(\%)$ and –7 % in $\Delta R_{95}(\%)$) at a lower capital cost (50×1000 USD). Together, these results suggest that distributed microgrid expansion offers the most cost-efficient resilience pathway, whereas proactive cooling-center activation provides the most operationally flexible short-term mitigation option.

**Table 4.** Parameterization of intervention scenarios in equity-adjusted risk simulation

| Intervention ID | Name / Description | Primary Target | Modeled Adjustments | Staff Hours per Day | Estimated Cost (× 1000 USD) | Description |
|---|---|---|---|---|---|---|
| I1 | Preemptive Cooling Center | Exposure, Vulnerability, Sensitivity | Exposure ↓ 30% (× 0.70) $\Delta V_{rank} = -0.05$ $\Delta E_{rank} = -0.10$ | 8 | 50 | Early activation of public shelters before onset of heat or outage conditions |
| I2 | Reactive Opening | Exposure, Sensitivity | Exposure ↓ 10% (× 0.90) $\Delta E_{rank} = -0.05$ | 3 | 15 | Delayed shelter operation after hazard thresholds are exceeded. |
| I3 | Microgrid (Clinic-First) | Exposure, Vulnerability | Exposure × 0.50 $\Delta V_{rank} = -0.05$ | 0 | 120 | Local backup power for healthcare facilities to ensure continuity of critical services. |
| I4 | Microgrid (Vulnerable Block) | Exposure | Exposure ↓ 35% (× 0.65) | 0 | 180 | Neighborhood-scale expansion of microgrid coverage to the top 15 % of high-risk buildings. |
| I5 | Targeted Outreach / Retrofits | Vulnerability, Sensitivity | $\Delta V_{rank} = -0.03$ $\Delta E_{rank} = -0.12$ | 6 | 30 | Community programs and household retrofits to enhance social and thermal resilience. |

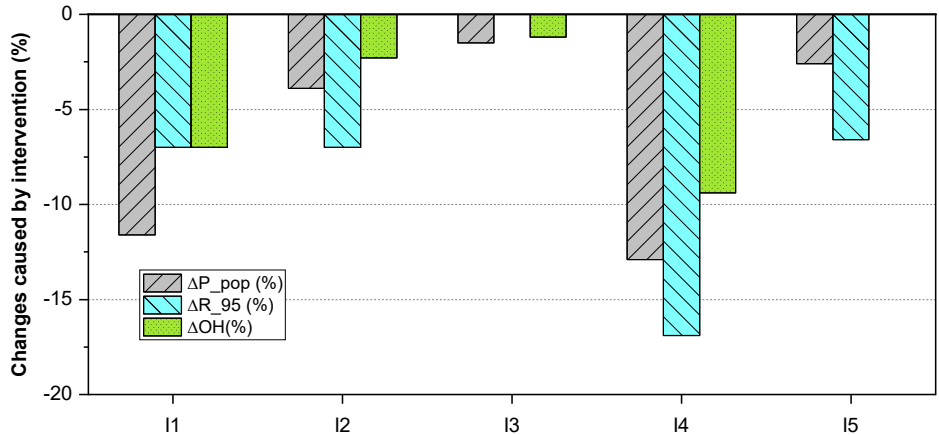

**Figure 9.** Equity-adjusted risk reduction for the adopted intervention scenarios

## 4. Research Challenges and Future Directions

While the H-RDT presents a comprehensive framework for urban-scale disaster resilience, transforming it into a fully operational system requires overcoming substantial scientific, computational, and ethical challenges. The integration of physics-based modeling, learning-driven affiliation dynamics, and human-centered decision processes creates a uniquely multidimensional problem. Models must remain physically consistent and numerically stable while operating on incomplete, uncertain data streams, and their outputs must also be socially interpretable and ethically defensible. Achieving this synthesis of physics, data, and human judgment represents one of the most pressing frontiers in digital-twin research.

A primary research priority lies in advancing hybrid model calibration and real-time data assimilation. Physics-informed neural networks (PINNs) and hybrid surrogates offer promising routes, yet their reliance on high-fidelity data is challenged by the sparse, delayed, or degraded observations characteristic of disaster scenarios. Adaptive calibration schemes that combine state-space updating (e.g. Kalman filtering or ensemble-based methods) with physics-constrained loss terms are needed so that H-RDT can continuously adjust boundary conditions, material properties, or forcing parameters. Efforts in online calibration of hybrid systems (e.g. Euler

Gradient Approximation, EGA) demonstrate how neural corrections can be updated in tandem with numerical solvers, even when the model is not fully differentiable (Ouala et al., 2024). Other recent work on hybrid semi-parametric models and adaptive weighting for Bayesian PINNs further illustrates the potential for balancing multiple objectives across scales and tasks (Kasilingam et al., 2024).

Equally critical is achieving scalability for compounding hazard simulation. Simultaneously representing multiple hazards, such as heat, outage, and flooding, requires computational architectures capable of handling thousands of spatial nodes and dynamically evolving boundary conditions. Research should focus on surrogate and reduced-order models that project high-dimensional physical systems into compact latent spaces, enabling rapid simulation under shifting environmental and infrastructural conditions. Multi-fidelity learning frameworks linking detailed building-level thermodynamic models with regional hazard propagation models can preserve essential accuracy while improving computational tractability. Incorporating uncertainty quantification directly into these frameworks is equally vital, as model parameters evolve with context and time. Embedding probabilistic inference and reinforcement-driven exploration would allow both epistemic and aleatory uncertainties to be represented explicitly, strengthening confidence in real-time decision outputs.

Another emerging direction involves advancing graph-based reinforcement learning for adaptive network topologies. The H-RDT must recognize changing interdependence among assets, such as how transportation, energy, and communication systems dynamically influence one another during disruption. However, maintaining stability and physical realism in such evolving graph structures remains challenging. Future algorithms should incorporate physics-constrained graph reinforcement learning, with updates to connectivity respect infrastructure hierarchies and conservation principles. Distributed learning across edge devices would further decentralize computation, allowing local agents to continue adapting even when connectivity to the central system is lost during a disaster.

Integrating social behavior and equity dynamics introduces a parallel set of challenges. Although the equity-aware risk index captures how socio-economic factors amplify vulnerability, it remains a static representation of inherently nonlinear human responses. Future research should embed behavioral models, such as agent-based representations of household adaptation, migration, or evacuation, into the decision layer so that resilience measures account for both infrastructural recovery and human adaptation. Coupling these behavioral modules with privacy-preserving analytics and participatory sensing will be critical for building public trust. Federated-learning architectures, which allow communities to retain ownership of their data while contributing model updates, provide a viable pathway toward ethical, inclusive learning frameworks that respect local autonomy (e.g., federated XAI) (Lopez-Ramos et al., 2024).

On the data and interoperability front, building H-RDT as a shared operational platform demands standardization in data protocols, metadata schemas, and ontologies. Aligning digital twin data structures with established resilience systems, such as FEMA's Hazus, to facilitate coordinated responses and institutional collaboration. Developing open APIs and standardized metadata, as well as adhering to ontologies for hazard, infrastructure, and social dimensions, will promote reproducibility, module sharing, and cross-study validation. Reviews of digital twin–AI integration highlight these challenges of interoperability and data fusion as central bottlenecks in practice (Hao et al., 2024; Kreuzer et al., 2024).

Beyond technical hurdles, ethical and governance issues must remain central. Integrating continuous IoT and social data streams introduces risks around privacy, consent, bias, and fairness. Uneven sensor coverage can distort the perceived equity landscape, potentially reinforcing the very disparities the model aims to mitigate. Embedding responsible-AI principles, such as explainability, uncertainty disclosure, auditability, and community oversight, across all stages of the system is essential. Pilot deployments in municipalities, utilities, or NGOs should evaluate both algorithmic performance and downstream implications of fairness, legitimacy, and accountability.

It is also important to note the limitations and transferability of the current study. We employed a synthetic urban district to enable controlled and repeatable analysis of compounding hazards and sensing loss. The indoor thermal model adopts a simplified 2R2C structure with a smoke-modulated ventilation factor; while appropriate for city-scale screening, it omits detailed envelope heterogeneity, humidity effects, and adaptive occupant behaviors. Reported benefits therefore represent conservative, first-order estimates. For real-city deployment, the H-RDT would require (i) building-archetype libraries or limited field audits, (ii) basic IoT data streams (zone temperatures, setpoints) or smart-meter proxies, and (iii) scheduled UAV or remote-sensing overpasses during events. The fusion, equity, and intervention modules can be transferred directly, with only model coefficients and priors re-calibrated through the same PINN–assimilation loop. Future validation using municipal datasets and targeted field pilots will be essential to demonstrate operational readiness and generalizability.

In summary, operationalizing H-RDT is challenging but intellectually rich. Progress will depend on bridging hybrid learning and numerical physics, scaling to multi-hazard complexity, embedding adaptive infrastructure–behavior coupling, standardizing across institutional systems, and incorporating ethics by design. Advancing these directions will establish digital twins not merely as diagnostic tools but as trusted, equitable decision-support systems for resilience in the era of compound urban hazards.

## 5. Conclusions

The H-RDT framework provides a novel pathway for integrating physics-based modeling, adaptive data fusion, and equity-centered analytics to enhance disaster resilience in urban environments. Applied to a synthetic urban district, the framework demonstrated its capability to couple sensing, simulation, and intervention under complex, multi-hazard conditions. The main findings from the case study are as follows:

- The physics-informed neural network (PINN) using a 2R2C configuration accurately reproduced indoor temperature evolution during heatwave–outage events, capturing peak temperatures of approximately 31–33°C and recovery behavior following power restoration. The hybrid model remained stable and accurate even under partial sensor loss, confirming its robustness for real-time hazard simulation.

- The reinforcement learning-based fusion module dynamically reweighted IoT, UAV, and satellite data to sustain continuous coverage. When ground sensors failed during outages, the algorithm automatically increased reliance on remote-sensing inputs, minimizing estimation error and preserving spatiotemporal continuity in the reconstructed temperature fields.

- Equity-adjusted risk analysis revealed the high-vulnerability groups, particularly school, clinics, and low-income housing, experienced disproportionately higher cumulative exposure. Decile-based analysis (D1–D10) quantified these disparities, showing that the upper deciles contributed most of the total risk burden despite representing a smaller share of population.

- Among the five interventions tested, Preemptive Cooling Center (I1) and Microgrid–Vulnerable Block (I4) strategies achieved the largest overall gains, lowering population-weighted risk by 11–13%, tail risk by 7–17%, and overheating hours by 7–9%. Reactive Opening (I2) provided moderate improvement (~4%, 7%, and 2%, respectively), while Targeted Outreach (I5) and Microgrid–Clinic-First (I3) yielded smaller but focused benefits (1–3%) for vulnerable households and critical facilities. Preemptive actions delivered broader, system-wide resilience, whereas reactive or localized measures produced targeted yet limited effects.

The results highlight two immediate, actionable practices: (1) anticipatory activation of cooling centers based on combined heat and grid-stress forecasts, and (2) microgrid expansion prioritized for top-decile risk clusters. Agencies can use the metrics $\Delta P_{pop}(\%)$ and $\Delta R_{95}(\%)$ to establish activation thresholds and evaluate cost-normalized trade-offs (e.g., staffing versus installed capacity). The decile analysis offers a transparent framework for equity-first sitting of shelters, distributed energy resources, and outreach programs, while the overheating-hour metric provides an intuitive indicator of health-related thermal risk. Overall, the H-RDT operates as a living, adaptive digital ecosystem that links physical hazard dynamics with social vulnerability and decision support. By uniting real-time sensing, hybrid physics–AI modeling, and equity-aware analytics, the framework moves beyond static assessments toward continuous, data-driven optimization of resilience in climate-stressed urban systems.

**Disclaimer**

*This paper has been authored by UT-Battelle, LLC, under contract DE-AC05-00OR22725 with the US Department of Energy (DOE). The US Government retains and the publisher, by accepting the article for publication, acknowledges that the US government retains a nonexclusive, paid-up, irrevocable, worldwide license to publish or reproduce the published form of this manuscript, or allow others to do so, for US government purposes. DOE will provide public access to these results of federally sponsored research in accordance with the DOE Public Access Plan ([http://energy.gov/downloads/doe-public-access-plan](http://energy.gov/downloads/doe-public-access-plan)).*

# References

Ariyachandra, M. R. M. F., & Wedawatta, G. (2023). Digital Twin Smart Cities for Disaster Risk Management: A Review of Evolving Concepts. *Sustainability*.

Booth, S., Reilly, J., Butt, R., Wasco, M., Monohan, R., Booth, S., Reilly, J., Butt, R., Wasco, M., & Monohan, R. (2019). *Microgrids for Energy Resilience: A Guide to Conceptual Design and Lessons from Defense Projects Microgrids for Energy Resilience* (Issue May).

Braik, A. M., & Koliou, M. (2024). A digital twin framework for efficient electric power restoration and resilient recovery in the aftermath of hurricanes considering the interdependencies with road network and essential facilities. *Resilient Cities and Structures*, *3*(July), 79–91. https://doi.org/10.1016/j.rcns.2024.07.004

Bruneau, M., Chang, S. E., Eguchi, R. T., Lee, G. C., O'Rourke, T. D., Reinhorn, A. M., Shinozuka, M., Tierney, K., Wallace, W. A., & Von Winterfeldt, D. (2003). A Framework to Quantitatively Assess and Enhance the Seismic Resilience of Communities. *Earthquake Spectra*, *19*(4), 733–752. https://doi.org/10.1193/1.1623497

Buldyrev, S. V., Parshani, R., Paul, G., Stanley, H. E., & Havlin, S. (2010). Catastrophic cascade of failures in interdependent networks. *Nature*, *464*(7291), 1025–1028. https://doi.org/10.1038/nature08932

Cimellaro, G. P., Renschler, C., Reinhorn, A. M., & Arendt, L. (2016). PEOPLES: A Framework for Evaluating Resilience. *Journal of Structural Engineering*, *142*(10). https://doi.org/10.1061/(asce)st.1943-541x.0001514

Cui, Y., Li, Y., Tang, H., Turowski, J. M., Yan, Y., & Bazai, N. A. (2024). A digital-twin platform for cryospheric disaster warning. *Natural Science Review*, *11*(inwae300).

Deren, L., Wenbo, Y., & Zhenfeng, S. (2021). Smart city based on digital twins. *Computational Urban Science*, *1*(1), 1–11. https://doi.org/10.1007/s43762-021-00005-y

Drysdale, D. (2011). An Introduction to Fire Dynamics. In *Wiley* (Issue July). https://doi.org/10.1002/9781119975465

Errandonea, I., Beltrán, S., & Arrizabalaga, S. (2020). *Computers in Industry Digital Twin for maintenance : A literature review*. *123*. https://doi.org/10.1016/j.compind.2020.103316

Fan, C., Zhang, C., Yahja, A., & Mostafavi, A. (2021). Disaster City Digital Twin: A vision for integrating artificial and human intelligence for disaster management. *International Journal of Information Management*, *56*(December 2019), 102049. https://doi.org/10.1016/j.ijinfomgt.2019.102049

Fogli, D. (2019). A Survey on Digital Twin: Definitions, Characteristics, Applications, and Design Implications. *IEEE Access*, *7*(Ml), 167653–167671. https://doi.org/10.1109/ACCESS.2019.2953499

Fu, N., Kim, M. K., Huang, L., Liu, J., Chen, B., & Sharples, S. (2022). Experimental and numerical analysis of indoor air quality affected by outdoor air particulate levels (PM1.0, PM2.5 and PM10), room infiltration rate, and occupants' behaviour. *Science of the Total*

*Environment*, *851*(May), 158026. https://doi.org/10.1016/j.scitotenv.2022.158026

Fuller, A., Member, S., & Fan, Z. (2020). Digital Twin: Enabling Technologies, Challenges and Open Research. *IEEE Access*, *8*. https://doi.org/10.1109/ACCESS.2020.2998358

Geng, Z., Zhang, C., Jiang, Y., Pugliese, D., & Cheng, M. (2025). Integrating multi-source data for life-cycle risk assessment of bridge networks: a system digital twin framework. *Journal of Infrastructure Preservation and Resilience*. https://doi.org/10.1186/s43065-025-00121-7

Ghaffarian, S. (2025). Rethinking digital twin: Introducing digital risk twin for disaster risk management. *Natural Hazards*, 1–5. https://doi.org/10.1038/s44304-025-00135-x

Gill, J. C., & Malamud, B. D. (2014). Reviews of Geophysics. *Reviews of Geophysics*, *52*, 680–722. https://doi.org/10.1029/88EO01108

Goodchild, M. F. (2007). Citizens as sensors: The world of volunteered geography. *GeoJournal*, *69*(4), 211–221. https://doi.org/10.1007/s10708-007-9111-y

Grieves, M., & Vickers, J. (2018). Digital Twin: Mitigating Unpredictable, Undesirable Emergent Behavior in Complex Systems. In *Transdisciplinary Perspective on Complex Systems* (Issue March). Springer. https://doi.org/10.1007/978-3-319-38756-7

Ham, Y., & Kum, J. (2020). Participatory sensing and digital twin city: Updating virtual city models for enhanced risk-informed decision-making. *Journal of Management in Engineering*, *36*(3).

Hamdanieh, L, Stephens, C., Olyaeemanesh, A., & Ostadtaghizadeh, A. (2024). Social justice: The unseen key pillar in disaster risk management. *International Journal of Disaster Risk Reduction*, *101*(December 2022), 104229. https://doi.org/10.1016/j.ijdrr.2023.104229

Hamidieh, M., & Ghassemi, M. (2022). Microgrids and Resilience: A Review. *IEEE Access*, *10*(September), 106059–106080. https://doi.org/10.1109/ACCESS.2022.3211511

Hao, N., Li, Y., Liu, K., Liu, S., Lu, Y., Xu, B., Li, C., Chen, J., Yue, L., Fu, T., Hu, X., Wang, X., & Zhao, Y. (2024). Artificial Intelligence-Aided Digital Twin Design: A Systematic Review. *Preprints.Org*, 202408.2063.v1. https://doi.org/10.20944/preprints202408.2063.v1

Hawe, G. I., Coates, G., Wilson, D. T., & Crouch, R. S. (2012). Agent-based simulation for large-scale emergency response: A survey of usage and implementation. *ACM Computing Surveys*, *45*(1). https://doi.org/10.1145/2379776.2379784

He, Z., & Weng, W. (2021). A Risk Assessment Method for Multi-Hazard Coupling Disasters. *Risk Analysis*, *41*(8), 1362–1375. https://doi.org/10.1111/risa.13628

Hondula, D. M., Davis, R. E., Saha, M. V., Wegner, C. R., & Veazey, L. M. (2015). Geographic dimensions of heat-related mortality in seven U.S. cities. *Environmental Research*, *138*, 439–452. https://doi.org/10.1016/j.envres.2015.02.033

Karniadakis, G. E., Kevrekidis, I. G., Lu, L., Perdikaris, P., Wang, S., & Yang, L. (2021). Physics- informed machine learning. *Nature Reviews Physics*, *3*(June). https://doi.org/10.1038/s42254-021-00314-5

Kasilingam, S., Yang, R., Singh, S. K., Farahani, M. A., Rai, R., & Wuest, T. (2024). Physics-

based and data-driven hybrid modeling in manufacturing: a review. *Production and Manufacturing Research*, *12*(1). https://doi.org/10.1080/21693277.2024.2305358

Kreuzer, T., Papapetrou, P., & Zdravkovic, J. (2024). Artificial intelligence in digital twins—A systematic literature review. *Data and Knowledge Engineering*, *151*(March), 102304. https://doi.org/10.1016/j.datak.2024.102304

Kritzinger, W., Karner, M., Traar, G., Henjes, J., & Sihn, W. (2018). Digital Twin in manufacturing: A categorical literature review and classification. *IFAC-PapersOnLine*, *51*(11), 1016–1022. https://doi.org/10.1016/j.ifacol.2018.08.474

Kryvasheyeu, Y., Chen, H., Obradovich, N., Moro, E., Van Hentenryck, P., Fowler, J., & Cebrian, M. (2016). Rapid assessment of disaster damage using social media activity. *Science Advances*, *2*(3), 1–11. https://doi.org/10.1126/sciadv.1500779

Lagap, U., & Ghaffarian, S. (2024a). Digital post-disaster risk management twinning: A review and improved conceptual framework. *International Journal of Disaster Risk Reduction*, *110*(June), 104629. https://doi.org/10.1016/j.ijdrr.2024.104629

Lagap, U., & Ghaffarian, S. (2024b). Digital post-disaster risk management twinning: A review and improved conceptual framework. *International Journal of Disaster Risk Reduction*, *110*(December 2023), 104629. https://doi.org/10.1016/j.ijdrr.2024.104629

Lemaitre, J. (1985). A continuous damage mechanics model for ductile fracture. *Journal of Engineering Materials and Technology, Transactions of the ASME*, *107*(1), 83–89. https://doi.org/10.1115/1.3225775

Liang, H., Moya, B., Seah, E., Ng, A., Weng, K., Baillargeat, D., Joerin, J., Zhang, X., Chinesta, F., & Chatzi, E. (2025). Harnessing hybrid digital twinning for decision-support in smart infrastructures. *Data-Centric Engineering*, *6*(e43). https://doi.org/10.1017/dce.2025.10015

Liang, Y., Sengupta, D., Campmier, M. J., Lunderberg, D. M., Apte, J. S., & Goldstein, A. H. (2021). Wildfire smoke impacts on indoor air quality assessed using crowdsourced data in California. *Proceedings of the National Academy of Sciences of the United States of America*, *118*(36), 1–6. https://doi.org/10.1073/pnas.2106478118

Liu, T., & Meidani, H. (2024). Graph Neural Network Surrogate for Seismic Reliability Analysis of Highway Bridge Systems. *Journal of Infrastructure Systemsucture System*, *30*(4).

Lopez-Ramos, L. M., Leiser, F., Rastogi, A., Hicks, S., Strümke, I., Madai, V. I., Budig, T., Sunyaev, A., & Hilbert, A. (2024). Interplay between Federated Learning and Explainable Artificial Intelligence: a Scoping Review. *ArXiv Preprint*, 1–16. http://arxiv.org/abs/2411.05874

Martini, T., Boigk, M., Catal, F., Dietze, S., Gerold, M., Lukau, E., Monteforte, M., Neuhäuser, S., Peitzsch, S., Phung, W., Pfennigschmidt, S., Simon, M., Vetter, J. Z., Winter, N., Adams, G., Finger, J., & Rosin, J. (2025). Digital twin representation of socio-technical systems through a distributed co-simulation approach for crisis management. *Environment Systems and Decisions*, *45*(3), 1–21. https://doi.org/10.1007/s10669-025-10035-0

Matthiss, B., Azzam, A., & Binder, J. (2023). Thermal Building Models for Energy Management Systems. *Proceedings - 2023 IEEE International Conference on Environment and*

*Electrical Engineering and 2023 IEEE Industrial and Commercial Power Systems Europe, EEEIC / I and CPS Europe 2023*, 1–6. https://doi.org/10.1109/EEEIC/ICPSEurope57605.2023.10194636

National Academies of Sciences, Engineering, and Medicine. (2024). *Compounding Disasters in Gulf Coast Communities, 2020-2021: Impacts, Findings, and Lessons Learned*. National Academies Press. https://doi.org/10.17226/27170

Ouala, S., Chapron, B., Collard, F., Gaultier, L., & Fablet, R. (2024). Online calibration of deep learning sub-models for hybrid numerical modeling systems. *Communications Physics*, *7*(1). https://doi.org/10.1038/s42005-024-01880-7

Ouyang, M. (2014). Review on modeling and simulation of interdependent critical infrastructure systems. *Reliability Engineering and System Safety*, *121*, 43–60. https://doi.org/10.1016/j.ress.2013.06.040

Panteli, M., & Mancarella, P. (2017). Modeling and evaluating the resilience of critical electrical power infrastructure to extreme weather events. *IEEE Systems Journal*, *11*(3), 1733–1742. https://doi.org/10.1109/JSYST.2015.2389272

Petkova, E. P., Vink, J. K., Horton, R. M., Gasparrini, A., Bader, D. A., Francis, J. D., & Kinney, P. L. (2017). Towards more comprehensive projections of urban heat-related mortality: Estimates for New York city under multiple population, adaptation, and climate scenarios. *Environmental Health Perspectives*, *125*(1), 47–55. https://doi.org/10.1289/EHP166

Pignatta, G., & Alibrandi, U. (2022). Risk-Informed Digital Twin (RDT) for the Decarbonization of the Built Environment: The Australian Residential Context. *Environmental Sciences Proceedings*, *12*(10).

Raissi, M., Perdikaris, P., & Karniadakis, G. E. (2019). Physics-informed neural networks : A deep learning framework for solving forward and inverse problems involving nonlinear partial differential equations. *Journal of Computational Physics*, *378*, 686–707. https://doi.org/10.1016/j.jcp.2018.10.045

Reames, T. G. (2016). A community-based approach to low-income residential energy efficiency participation barriers. *Local Environment*, *21*(12), 1449–1466. https://doi.org/10.1080/13549839.2015.1136995

Rinaldi, S. M., Peerenboom, J. P., & Kelly, T. K. (2001). Identifying, understanding, and analyzing critical infrastructure interdependencies. *IEEE Control Systems Magazine*, *21*(6), 11–25. https://doi.org/10.1109/37.969131

Rohat, G., Wilhelmi, O., Flacke, J., Monaghan, A., Gao, J., van Maarseveen, M., & Dao, H. (2021). Assessing urban heat-related adaptation strategies under multiple futures for a major U.S. city. *Climatic Change*, *164*(3–4). https://doi.org/10.1007/s10584-021-02990-9

Shen, Z., Chen, C., Fefferman, N., Zhou, H., & Shrestha, S. (2022). Community Vulnerability is the Key Determinant of Diverse Energy Burdens in The United States. *Energy Research & Social Science*.

Soden, R., Lallemant, D., Kalirai, M., Liu, C., Wagenaar, D., & Jit, S. (2023). The importance of accounting for equity in disaster risk models. *Communications Earth and Environment*, *4*(1),

1–8. https://doi.org/10.1038/s43247-023-01039-2

Sun, L., D'Ayala, D., Fayjaloun, R., & Gehl, P. (2021). Agent-based model on resilience-oriented rapid responses of road networks under seismic hazard. *Reliability Engineering and System Safety*, *216*(April), 108030. https://doi.org/10.1016/j.ress.2021.108030

Tao, F., Member, S., Zhang, H., Liu, A., & Nee, A. Y. C. (2019). Digital Twin in Industry: State-of-the-Art. *IEEE Transactions on Industrial Informatics*, *15*(4), 2405–2415. https://doi.org/10.1109/TII.2018.2873186

Tezzele, M., Carr, S., Topcu, U., Willcox, K. E., & States, U. (2024). Adaptive planning for risk-aware predictive digital twins. *ArXiv Preprint*, 1–16.

Tilloy, A., Malamud, B. D., Winter, H., & Joly-Laugel, A. (2019). A review of quantification methodologies for multi-hazard interrelationships. *Earth-Science Reviews*, *196*(January), 102881. https://doi.org/10.1016/j.earscirev.2019.102881

Tonn, B., Carroll, D., Rose, E., Hawkins, B., Pigg, S., Bausch, D., Dalhoff, G., Blasnik, M., Eisenberg, J., & Cowan, C. (2014). Weatherization Works - Summary of Findings from the Retrospective Evaluation of the U.S. Department of Energy's Weatherization Assistance Program. In *Oak Ridge National Laboratory*.

U.S. EPA. (2016). Excessive Heat Events Guidebook. In *United States Environmental Protection Agency*.

Ugliotti, F. M., Osello, A., Daud, M., & Yilmaz, O. O. (2023). Enhancing Risk Analysis toward a Landscape Digital Twin Framework: A Multi-Hazard Approach in the Context of a Socio-Economic Perspective. *Sustainability*, *15*(12429). https://doi.org/10.3390/su151612429

Wang, Y., Yue, Q., Lu, X., Gu, D., Xu, Z., Tian, Y., & Zhang, S. (2024). Digital twin approach for enhancing urban resilience: A cycle between virtual space and the real world. *Resilient Cities and Structures*, *3*(2), 34–45. https://doi.org/10.1016/j.rcns.2024.06.002

Wu, Z., Pan, S., Chen, F., Long, G., Zhang, C., & Yu, P. S. (2021). A Comprehensive Survey on Graph Neural Networks. *IEEE Transactions on Neural Networks and Learning Systems*, *32*(1), 4–24. https://doi.org/10.1109/TNNLS.2020.2978386

Zhou, Jian, Zheng, W., & Wang, D. (2024). A resilient network recovery framework against cascading failures with deep graph learning. *Journal of Risk and Reliability*, *238*(1), 193–203. https://doi.org/10.1177/1748006X221128869

Zhou, Junjie. (2025). Multi-dimensional model and interactive simulation of intelligent construction based on digital twins. *Scientific Reports*, *15*(32189), 1–15.

Zhu, M., & Jin, J. (2025). Data-Driven Urban Digital Twins and Critical Infrastructure Under Climate Change: A Review of Frameworks and Applications. *Urban Planning*, *10*(10109).

Zio, E., & Miqueles, L. (2024). Digital twins in safety analysis, risk assessment and emergency management. *Reliability Engineering and System Safety*, *246*, 110040. https://doi.org/10.1016/j.ress.2024.110040

Zscheischler, J., Martius, O., Westra, S., Bevacqua, E., Raymond, C., Horton, R. M., van den

Hurk, B., AghaKouchak, A., Jézéquel, A., Mahecha, M. D., Maraun, D., Ramos, A. M., Ridder, N. N., Thiery, W., & Vignotto, E. (2020). A typology of compound weather and climate events. *Nature Reviews Earth and Environment*, *1*(7), 333–347. https://doi.org/10.1038/s43017-020-0060-z

# Appendix: Algorithms of Hazard-Responsive Digital Twin (H-RDT) Applied in Case Study

**Table A1**. PINN model

---
**Algorithm 1: PINN-2R2C: Training Workflow with Multimodal Observations**
---

**Input:** nodes, time series $\{t, T_{\text{out}}, u\}$, and multimodal data streams $\{\text{IoT, UAV, SAT}\}$
**Output:** Trained network parameters $\theta$, physics parameters $\phi$, and validation metrics

**Preprocessing:**
1. Construct time encodings $\sin(2\pi t/24)$, $\cos(2\pi t/24)$ and assign type IDs.
2. Merge multimodal observations with IoT $\to$ UAV $\to$ SAT priority.
3. Normalize inputs and outputs using global statistics $(\mu, \sigma)$.

**Model Definition:**
4. MLP with type embedding predicts initial states $(\widehat{T}_w^0, \widehat{T}_z^0)$.
5. Per-type physical parameters: $\phi = \{C_w, C_z, R_{wo}, R_{wz}, Q_{int}, Q_{max}, db\}$, enforced positive via exponential mapping.
6. Integrate 2R2C ODEs with Runge–Kutta (RK2) including smoke, solar, and outage logic.

**Window-Based Sequence Loss:**
7. For each node $i$ and time window $\mathcal{I}$, rollout $T_z^{\text{pred}}(t)$ via RK2.
8. Compare with available $T^{\text{obs}}(t)$ using Huber loss, weighted by $(1 + 0.5u(t))$ during outage.
9. Apply soft bounds during blackout: penalize $T_z < 31.5°C$ or $T_z > 35.5°C$.

**Auxiliary Physics Residual:**
10. Randomly sample points $x$; compute $\frac{\partial T}{\partial t}$ via autograd.
11. Evaluate residuals from 2R2C balance equations:
$R_w, R_z \to \mathcal{L}_{\text{phys}} = E[(R_w/S)^2 + (R_z/S)^2]$.

**Training Loop:**
12. For each epoch:
    (a) Sample $K$ windows across well-observed nodes.
    (b) Compute $\mathcal{L} = \mathcal{L}_{\text{seq}} + 12\,\mathcal{L}_{\text{phys}} + \mathcal{L}_{\text{soft}}$.
(c) Backpropagate, clip gradients, update with Adam, and adjust learning rate via scheduler.

**Validation and Export:**
13. Perform full-series rollouts for validation nodes.
14. Compute RMSE/MAE across IoT, UAV, and SAT sources.
15. Save figures, time series CSVs, and metric reports.
**return** $\theta$, $\phi$, validation metrics

---

**Table A2.** Adaptive multimodal data fusion

---

**Algorithm 1: Adaptive Multimodal Fusion: Weights for IoT, UAV, and SAT Streams**

---

**Input:** DATA_DIR, OUT_DIR; time series $T_{\text{out}}$ (length $T$); streams {IoT dense ($N \times T_{\text{iot}}$), UAV sparse ($N \times K_u$) with indices $I_u$, SAT sparse ($N \times K_s$) with indices $I_s$}; hyperparameters: window $W$, EMA factor $\beta$, softmax temperature $\tau$, freshness half-lives $\{\tau_u, \tau_s\}$

**Output:** Series of fusion weights $\mathbf{w}(t) = [w_{\text{iot}}, w_{\text{uav}}, w_{\text{sat}}]_{t=0}^{T-1}$; CSV/JSON/PNG diagnostics

**Configuration and loading:**
1. Set random seed and ensure figure/output folders exist.
2. Read `time.csv` to get authoritative length $T$; load `streams.npz`.
3. Require IoT; UAV and SAT are optional. Define global $T = \max(T, T_{\text{iot}}, \max(I_u)+1, \max(I_s)+1)$.

**Helper functions:**
4. `recent_window_dense`(array, $t$, $W$): return trailing window in column space for dense arrays.
5. `recent_window_sparse`(array, indices, $t$, $W$): map global time $t$ to last observed column and return trailing window.
6. `consistency_fixed`(window): map mean absolute finite difference to $[0,1]$ where higher is smoother.
7. `availability`(window): fraction of non NaN entries.
8. `freshness_strict`($t$, indices, $\tau$): $\exp(-\text{age}(t)/\tau)$ using steps since most recent observation.

**Fusion loop (score → softmax → EMA):**
9. Initialize $\mathbf{w}^{\text{EMA}} \leftarrow [1,1,1]/3$.
10. for $t = 0$ to $T-1$ do
    Build windows: $W_{\text{iot}} \leftarrow$ `recent_window_dense`, $W_{\text{uav}} \leftarrow$ `recent_window_sparse`, $W_{\text{sat}} \leftarrow$ `recent_window_sparse`.
    Compute freshness gates $f_{\text{uav}} =$ `freshness_strict`($t, I_u, \tau_u$) and $f_{\text{sat}} =$ `freshness_strict`($t, I_s, \tau_s$).
    Quality metrics: $r. =$ `consistency_fixed`($W.$), $q. =$ `availability`($W.$).
    Scores (availability emphasized):

    $s_{\text{iot}} = 0.8\, q_{\text{iot}} + 0.2\, r_{\text{iot}}, \quad s_{\text{uav}} = (0.8\, q_{\text{uav}} + 0.2\, r_{\text{uav}})\, f_{\text{uav}}, \quad s_{\text{sat}} = (0.8\, q_{\text{sat}} + 0.2\, r_{\text{sat}})\, f_{\text{sat}}.$

    Temperatured softmax: $\mathbf{w} = \text{softmax}((\mathbf{s} - \max \mathbf{s})/\tau)$.
    EMA smoothing and renormalization: $\mathbf{w}^{\text{EMA}} \leftarrow \beta\, \mathbf{w}^{\text{EMA}} + (1-\beta)\, \mathbf{w}$, then $\mathbf{w}^{\text{EMA}} \leftarrow \mathbf{w}^{\text{EMA}}/\|\mathbf{w}^{\text{EMA}}\|_1$.
    Append $\mathbf{w}^{\text{EMA}}$ to output series.

**Persistence and plotting:**
11. Save fusion series to OUT_DIR/fusion_weights_series.csv.
12. Save final weights and $T$ to OUT_DIR/fusion_weights.json.
13. Save plotting CSV to figures/fusion_weights.csv and plot $\mathbf{w}(t)$ to figures/fusion_weights.png.

**return** $\{\mathbf{w}(t)\}_{t=0}^{T-1}$ *and diagnostics.*

**Table A3.** GRL update with coverage-aware and edge-gain overlay

---
**Algorithm 1:** GRL Update with Coverage-Aware Criticals and Edge-Gain Overlay
---

**Input:** Node table with (id, $x$, $y$, type, pop, req); time series $\{\text{outage}(t), T_{\text{out}}(t)\}_{t=0}^{T-1}$; hyperparameters: $k$-NN degree ($k \leq 8$), GRL step $\eta$, decay $\gamma$, top-$M$ criticals per community $M$, animation flags/steps.

**Output:** Learned graph $H$ with updated edge weights and costs; critical nodes (per-community and global top-$K$); edge-gain figure(s) and optional GIF; JSON export.

**Data Load and Setup:**
1. Read `nodes.csv`, `time.csv`; if outage absent, set to zeros.
2. Initialize figure/output folders and random seed.

**Base Graph Construction (spatial $k$-NN):**
3. Create undirected graph $G$ with node attributes: type, position $(x, y)$, pop, req.
4. For each node, connect to $k$ nearest neighbors by Euclidean distance $d$; set initial edge weight $w_0 = \exp(-d/\sigma)$ and store ($w = w_0$, $w_0$, $d$).

**GRL-Style Edge Updates over Time:**
5. Build a normalized stress proxy $s(t) \in [0, 1]$ from outage and heat (if $T_{\text{out}}$ available).
6. For $t = 0 \ldots T - 1$:
   1. Node priors $p(n)$ from type (e.g., hospital>shelter>school>...).
   2. Node scores $u_t(n) = p(n) \cdot (1 + s(t))$; mean-center $u_t$.
   3. For each edge $(u, v)$: gradient $g_t = u_t(u) + u_t(v)$; update weight $w \leftarrow \gamma w + \eta g_t$, then clip to $[10^{-2}, 10]$.
   4. If $t$ is a snapshot index, store edge weights for animation.

**Learned Graph and Centralities:**
7. Set $H \leftarrow G$; define traversal cost on edges: $c(u, v) = 1/(w + \varepsilon)$.
8. Compute centralities on $H$: betweenness (cost-weighted), closeness (cost), eigenvector (weight).
9. Normalize each centrality and demographic proxies (pop, req); form composite:

$$s_{\text{cent}} = 0.5\,\hat{b} + 0.3\,\hat{c} + 0.2\,\hat{e}, \quad s_{\text{dem}} = (0.5 + 0.5\,\hat{\text{pop}})(0.5 + 0.5\,\hat{\text{req}}), \quad \text{crit\_score} = s_{\text{cent}} \cdot s_{\text{dem}}.$$

**Coverage-Aware Critical Selection:**
10. Detect communities via greedy modularity; for each community, select top-$M$ by `crit_score`.
11. Also compute global top-$K$ critical nodes for reporting.

**Edge-Gain Overlay and Animation (optional):**
12. For each edge, compute relative gain $\Delta = (w - w_0)/(w_0 + \varepsilon)$; color/thicken by $\Delta$ percentiles.
13. Plot overlay; if snapshots exist, render frames and (if available) assemble a GIF.
**return** $H$, critical sets, visualizations, and JSON report.

**Table A4.** Equity-adjusted risk

---
**Algorithm 1:** Equity-Adjusted Risk Computation and Visualization
---

**Input:** Node table with (id, $x$, $y$, pop, vuln, energy_burden, income, type); time series $\{outage(t), T_{out}(t)\}_{t=0}^{T-1}$; hyperparameters: $\Gamma$ (equity amplification), $\beta_{phys}$, $\beta_{sens}$, heat threshold $T_{th}$, exposure floor $\epsilon_{exp}$, smoothing neighborhood $K$.

**Output:** Per-node equity-adjusted risk $R_{node}$; aggregate community equity index $R_{eq}$; visualizations and summary tables (CSV, JSON, and PNG maps).

**Setup and I/O:**
1. Load node attributes from nodes.csv and time series from time.csv.
2. Fill missing outage column with zeros.
3. Define default vulnerability and energy-burden proxies if columns missing:
   - Vulnerability: type-based prior (hospital>shelter>school>grocery>generic).
   - Energy burden: directly from column; else invert normalized income; else rank of populati

**Helper Functions:**
4. percentile_rank(series): normalize to [0,1], return 0.5 if constant.
5. knn_smooth(xy, values, K): KNN average (includes self) using Euclidean distance.

**Exposure Calculation:**
6. If outdoor temperature available and $T \geq 12$:
   compute heat excess $h(t) = \max(T_{out}(t) - T_{th}, 0)$.
   normalize mean $\bar{h}$ to $[0,1]$ as heat_norm $= \min(\bar{h}/10, 1)$.
7. Compute outage fraction outage_frac $= \text{mean}(\text{outage} > 0.5)$.
8. Combine exposures:

$$\text{exposure}_{sys} = \max(0.5\,\text{outage\_frac} + 0.5\,\text{heat\_norm}, \epsilon_{exp}).$$

**Ranking and Spatial Smoothing:**
9. Rank-normalize $V$ (physical vulnerability) and $E$ (energy burden).
10. If $K > 1$, apply spatial KNN smoothing to $V$ and $E$, then re-rank.
11. Composite node term:

$$T_{node} = \beta_{phys}V + \beta_{sens}E.$$

If nearly constant, replace by ranked values.

**Risk Indices:**
12. Per-node risk:

$$R_{node} = \text{exposure}_{sys} \times T_{node}.$$

13. Population-weighted community index (Eq.(14))

**Outputs:**
14. Produce decile-based bar charts: unweighted and population-weighted mean risk.
**return** $R_{node}$, $R_{eq}$, CSV/JSON outputs, and spatial visualizations.

**Table A5.** Intervention over equity-adjusted risk

---
**Algorithm 1: Interventions over Equity-Adjusted Risk**
---

**Input:** `nodes.csv`, `time.csv`; optional: `equity_per_node.csv`, `equity_summary.json`. If equity files absent, synthesize defaults. Hyperparameters inherited: $\beta_{\text{phys}}$, $\beta_{\text{sens}}$ (from equity summary or defaults).

**Output:** `outputs/intervention_report.json`, `outputs/intervention_table.csv`; optional figure `figures/intervention_bars.png`.

**0) Load with graceful fallbacks**
1. Ensure output/figure directories exist.
2. Read `nodes.csv`, `time.csv`.
3. If `equity_per_node.csv` present, load and merge `type` if needed; else create minimal frame with $V = E = 0.5$, exposure $= 0.5$, and $R_{\text{node}} = 0.5\,(0.6 \cdot 0.5 + 0.4 \cdot 0.5)$.
4. If `equity_summary.json` present, read $\beta_{\text{phys}}$, $\beta_{\text{sens}}$; else set $(0.6, 0.4)$.

**1) Baseline metrics**
5. Infer total hours $H$ from `time.csv` (use $\Delta t$ from `time_h` if available; else 10 min).
6. System overheat hours $OH = \sum \mathbf{1}\{T_{\text{out}} > 30°C\}\,\Delta t$.
7. Let $P_i$ be node populations; compute baseline:

$$\bar{R} = \frac{\sum_i P_i\,R_{\text{node},i}}{\sum_i P_i}, \quad R_{95} = \text{percentile}_{95}(R_{\text{node}}).$$

**2) Intervention engine**
8. `apply_intervention(·)`: given masks and deltas/scales,

$$\text{exposure}' = \text{clip}(\text{exposure} \times s, 0, 1), \quad V' = \text{clip}(V + \Delta_V, 0, 1), \quad E' = \text{clip}(E + \Delta_E, 0, 1),$$

$$R'_{\text{node}} = \text{exposure}' \cdot (\beta_{\text{phys}} V' + \beta_{\text{sens}} E').$$

9. `eval_metrics(·)`: report deltas vs baseline (lower is better):

$$\Delta \bar{R}\% = 100 \frac{\bar{R}' - \bar{R}}{\bar{R} + 10^{-12}}, \quad \Delta R_{95}\%, \quad \Delta OH\% \text{ via mean exposure scaling.}$$

Include staffing hours/day and optional cost (kUSD).

**3) Define simple interventions**
10. Build masks: top decile by $R_{\text{node}}$; top 15% by $R_{\text{node}}$; clinics by type.
11. Instantiate examples:
    - I1 (preemptive cooling): scale exposure 0.70, $\Delta V = -0.05$, $\Delta E = -0.10$ on top decile.
    - I2 (reactive opening): scale exposure 0.90, $\Delta E = -0.05$ on top decile.
    - I3 (microgrid clinic-first): clinics exposure $\times 0.50$, $\Delta V = -0.05$.
    - I4 (microgrid share vulnerable): top 15% exposure $\times 0.65$.
    - I5 (targeted outreach/retrofit): top decile $\Delta V = -0.03$, $\Delta E = -0.12$.
12. Evaluate each: collect $\Delta \bar{R}\%$, $\Delta R_{95}\%$, $\Delta OH\%$, staffing/cost.

**4) Pareto filtering**
13. Non-dominated set by (improvement vs. resource): Staff Pareto: minimize staff_hours/day and $\Delta \bar{R}\%$.

**return** *Intervention outcomes, Pareto fronts, and saved artifacts.*